%% file: sample-acmtog-SIGGRAPH-submission.tex
\documentclass[acmtog]{acmart}

\usepackage{booktabs} 
\usepackage{graphicx}
\usepackage{makecell}
\usepackage{array}
\usepackage{subcaption}
\usepackage{wrapfig}
\usepackage{algpseudocode}
\usepackage{algorithm}
\usepackage{amsmath}
\usepackage{gensymb}
\usepackage{caption}
\usepackage{svg}
\usepackage{float}
\usepackage{multirow}

\renewcommand\footnotetextcopyrightpermission[1]{} 
\begin{document}
\title{ConTEXTure: Consistent Multiview Images to Texture}

\author{Jaehoon Ahn}
\affiliation{
 \institution{Sogang University}
 \country{South Korea}}
\email{jahn@sogang.ac.kr}
\author{Sumin Cho}
\affiliation{
 \institution{Sogang University}
 \country{South Korea}}
\email{suminyj@gmail.com}
\author{Harim Jung}
\affiliation{
 \institution{SwatchOn Co., Ltd}
 \country{South Korea}}
\email{jungharim225@gmail.com}
\author{Kibeom Hong}
\affiliation{
 \institution{Sookmyung Women's University}
 \country{South Korea}}
\email{kb.hong@sookmyung.ac.kr}
\author{Seonghoon Ban}
\affiliation{
 \institution{RECON Labs Inc.}
 \country{South Korea}}
\email{vincent@reconlabs.ai}
\author{Moon-Ryul Jung}
\affiliation{
 \institution{Sogang University}
 \country{South Korea}}
\email{moon@sogang.ac.kr}

\begin{abstract}
We introduce ConTEXTure, a generative network designed to create a texture map/atlas for a given 3D mesh using images from multiple viewpoints. The process begins with generating a front-view image from a text prompt, such as 'Napoleon, front view', describing the 3D mesh. Additional images from different viewpoints are derived from this front-view image and camera poses relative to it. ConTEXTure builds upon the TEXTure network, which uses text prompts for six viewpoints (e.g., 'Napoleon, front view', 'Napoleon, left view', etc.). However, TEXTure often generates images for non-front viewpoints that do not accurately represent those viewpoints.To address this issue, we employ Zero123++, which generates multiple view-consistent images for the six specified viewpoints simultaneously, conditioned on the initial front-view image and the depth maps of the mesh for the six viewpoints. By utilizing these view-consistent images, ConTEXTure learns the texture atlas from all viewpoint images concurrently, unlike previous methods that do so sequentially. This approach ensures that the rendered images from various viewpoints, including back, side, bottom, and top, are free from viewpoint irregularities.
\end{abstract}

%


%
%


\maketitle

\input{samplebody-journals}

\end{document}

%% file: samplebody-journals.tex
\captionsetup[subfloat]{justification=centering}

\begin{figure*}[!htbp]
    \centering

    \subfloat[``white humanoid robot, movie poster,\\villain character of a science fiction movie"]{
        \begin{minipage}{0.48\textwidth}
            \centering
            \includegraphics[trim={5cm 2cm 5cm 0},clip,width=0.32\textwidth,bb=0 0 724 724]{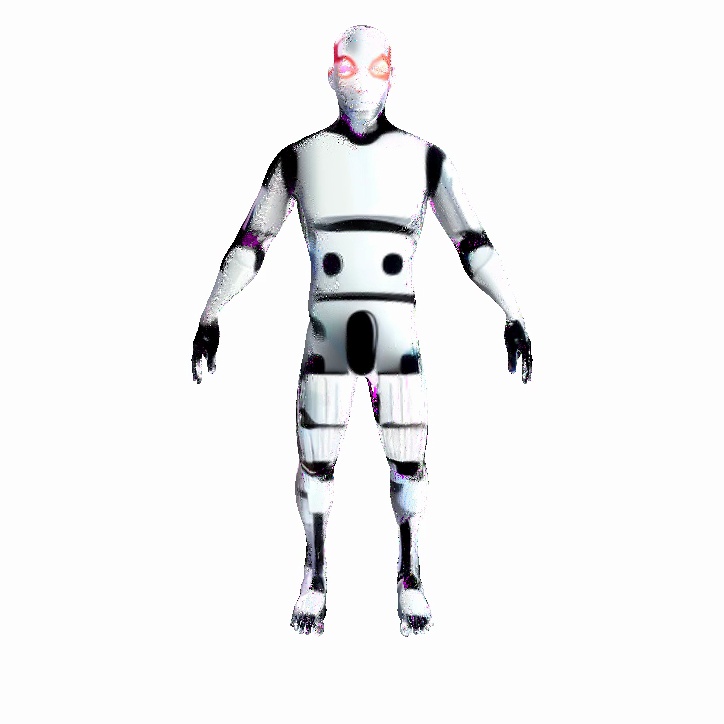}
            \includegraphics[trim={5cm 2cm 5cm 0},clip,width=0.32\textwidth,bb=0 0 724 724]{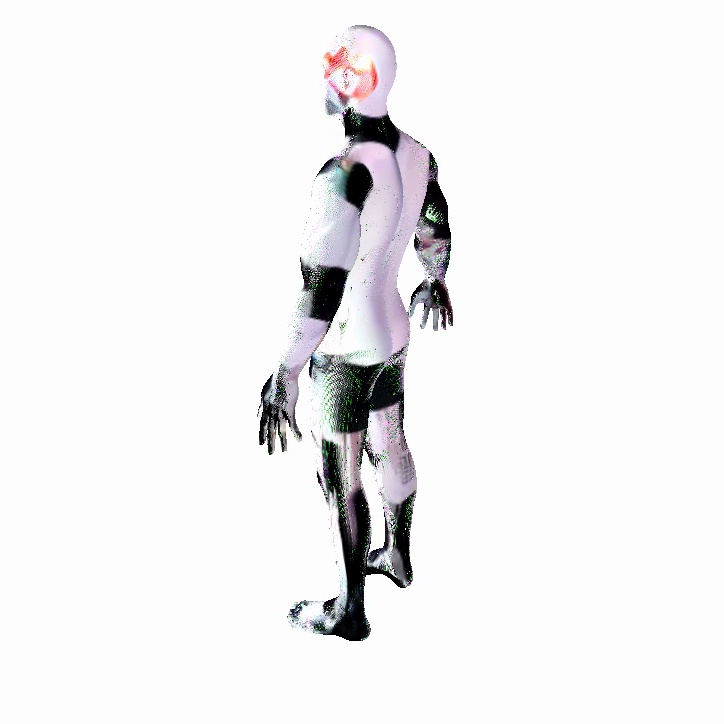}
            \includegraphics[trim={5cm 2cm 5cm 0},clip,width=0.32\textwidth,bb=0 0 724 724]{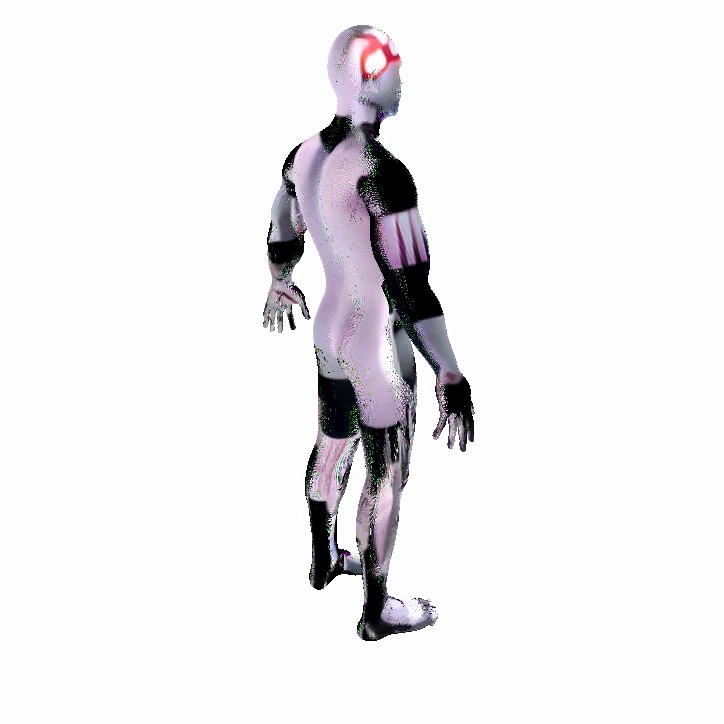}
        \end{minipage}
    }
    \hfill
    \subfloat[``doctor in a lab coat with a simple, modest hijab"]{
        \begin{minipage}{0.48\textwidth}
            \centering
            \includegraphics[trim={5cm 2cm 5cm 0},clip,width=0.32\textwidth,bb=0 0 724 724]{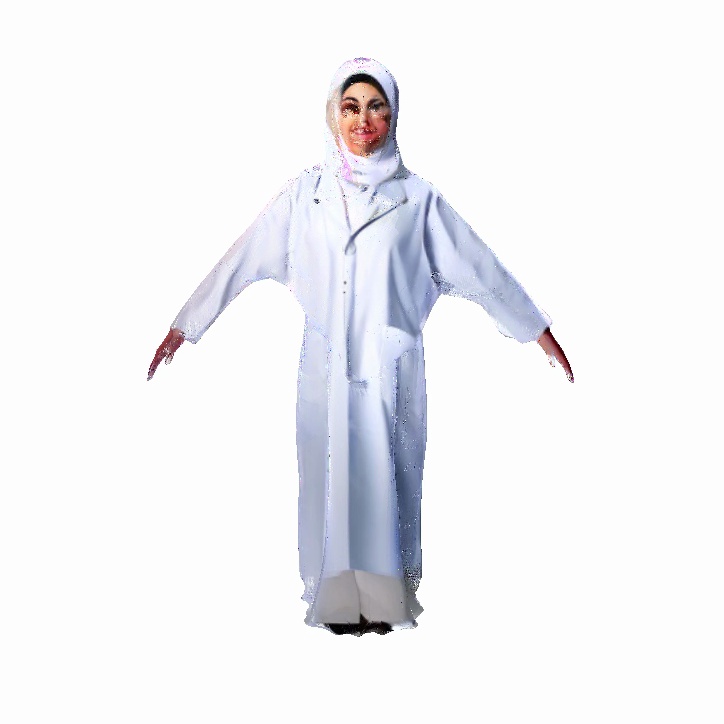}
            \includegraphics[trim={5cm 2cm 5cm 0},clip,width=0.32\textwidth,bb=0 0 724 724]{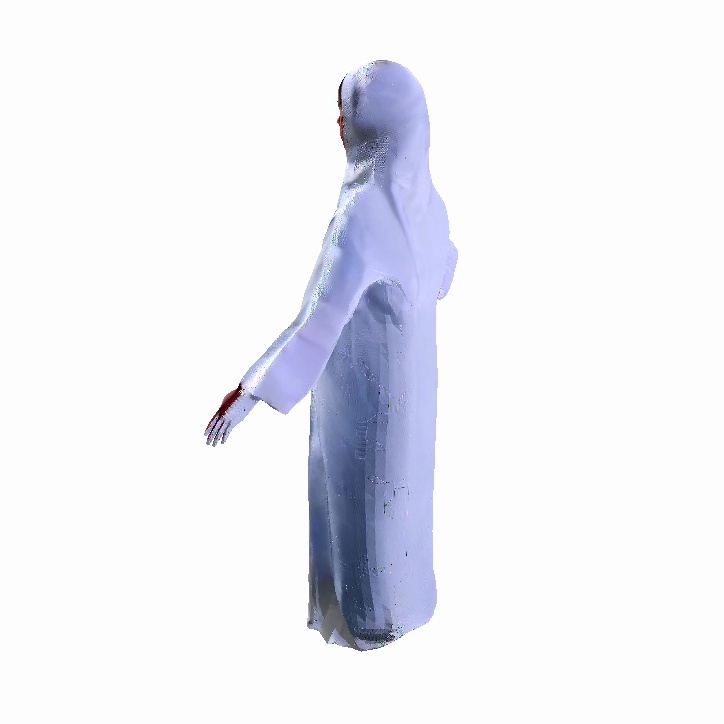}
            \includegraphics[trim={5cm 2cm 5cm 0},clip,width=0.32\textwidth,bb=0 0 724 724]{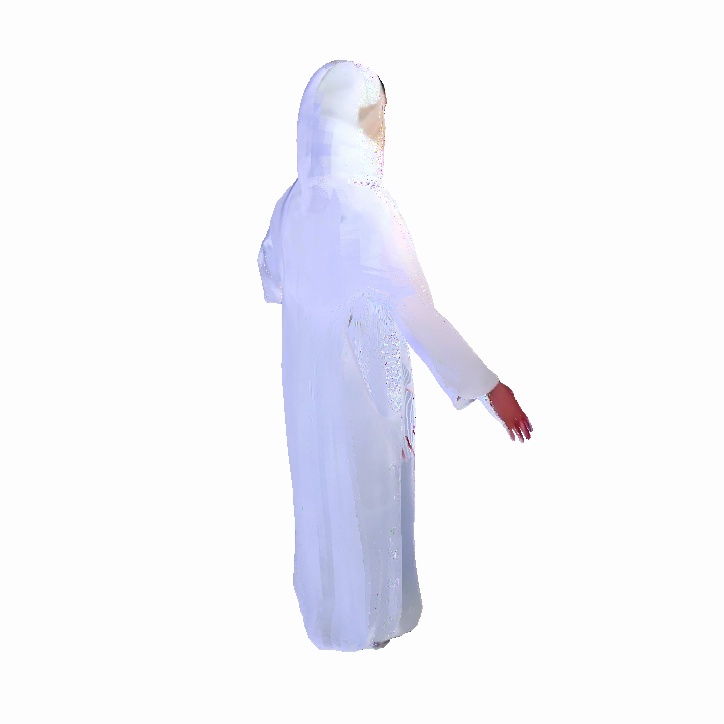}
        \end{minipage}
    }

    \vspace{0em} 

    \subfloat[``person in red sweater, blue jeans"]{
        \begin{minipage}{0.48\textwidth}
            \centering
            \includegraphics[trim={5cm 2cm 5cm 0},clip,width=0.32\textwidth,bb=0 0 724 724]{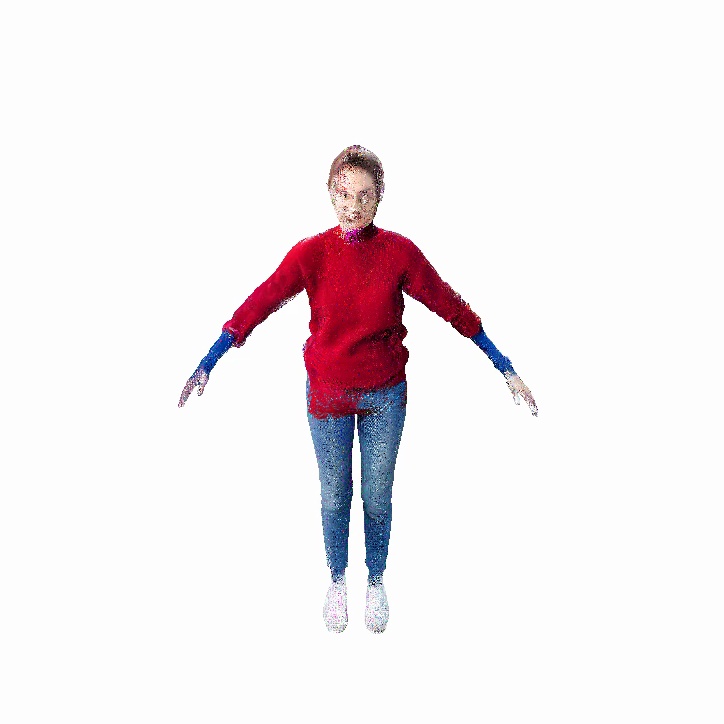}
            \includegraphics[trim={5cm 2cm 5cm 0},clip,width=0.32\textwidth,bb=0 0 724 724]{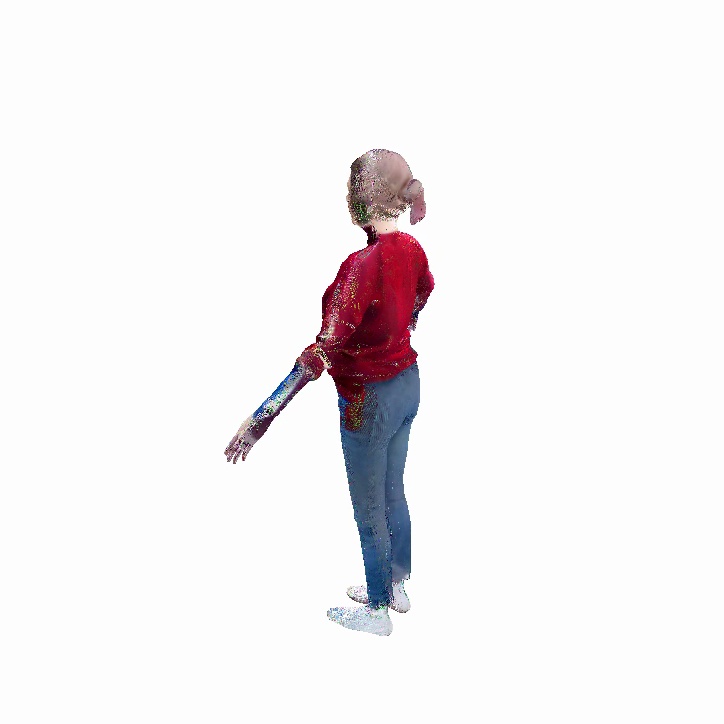}
            \includegraphics[trim={5cm 2cm 5cm 0},clip,width=0.32\textwidth,bb=0 0 724 724]{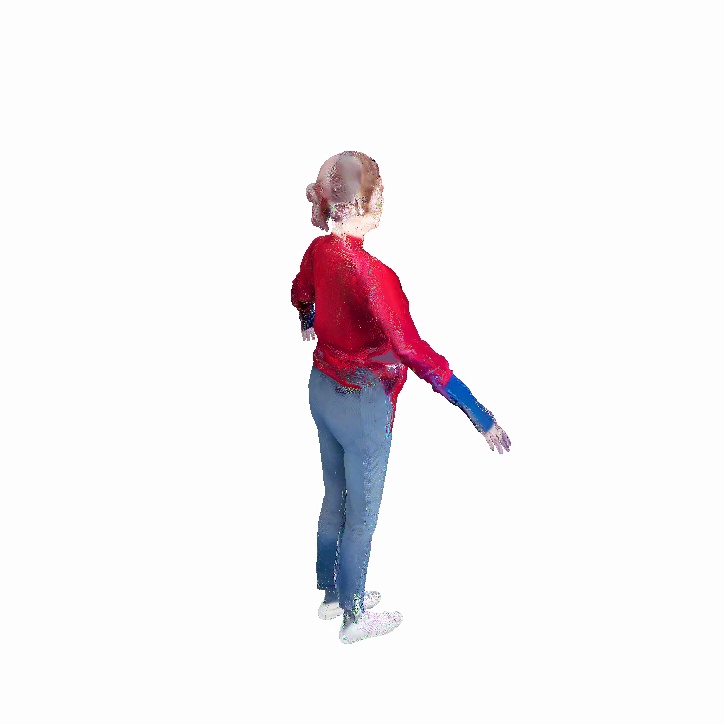}
        \end{minipage}
    }
    \hfill
    \subfloat[``rock star with leather jacket"]{
        \begin{minipage}{0.48\textwidth}
            \centering
            \includegraphics[trim={5cm 2cm 5cm 0},clip,width=0.32\textwidth,bb=0 0 724 724]{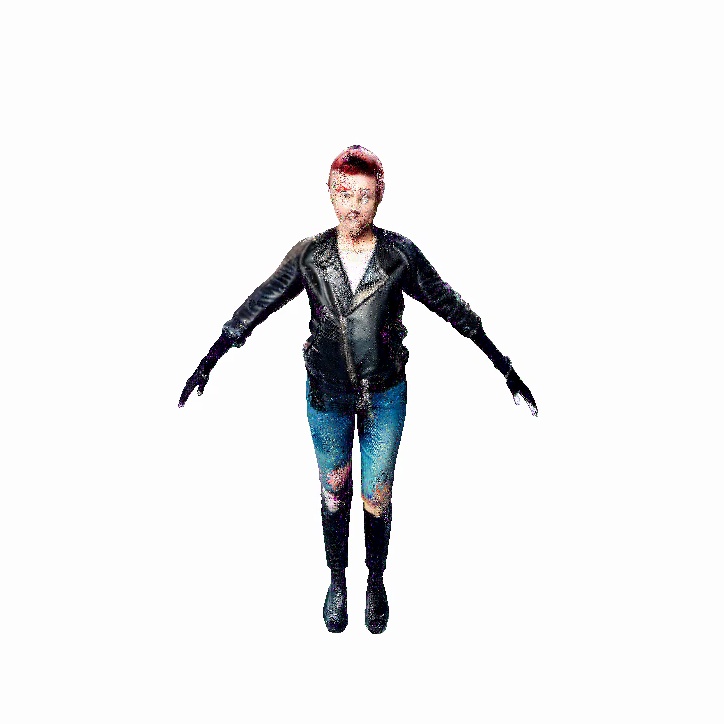}
            \includegraphics[trim={5cm 2cm 5cm 0},clip,width=0.32\textwidth,bb=0 0 724 724]{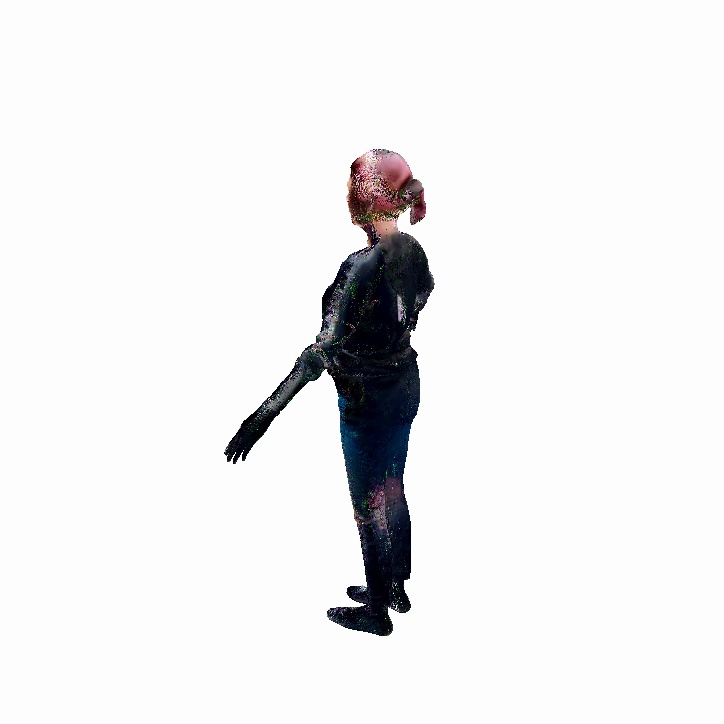}
            \includegraphics[trim={5cm 2cm 5cm 0},clip,width=0.32\textwidth,bb=0 0 724 724]{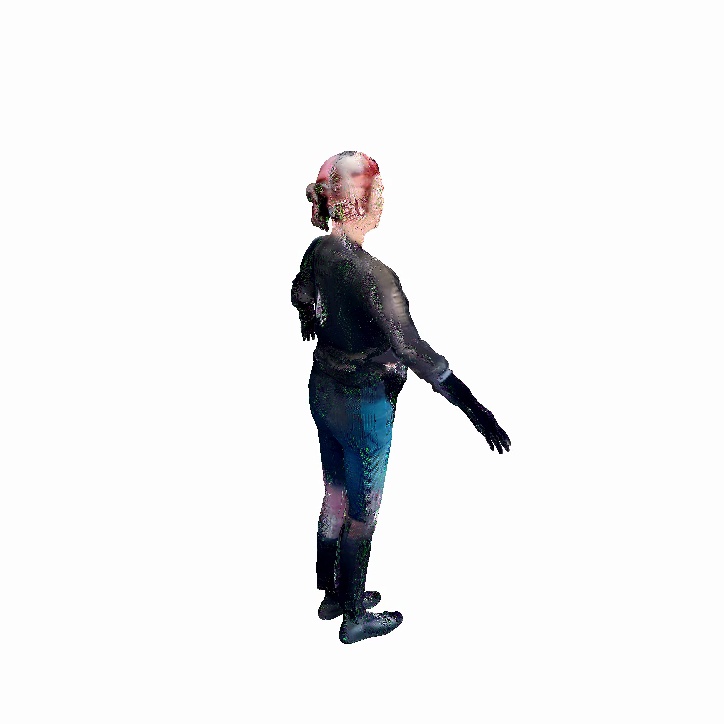}
        \end{minipage}
    }

    \caption{Four examples of texture maps generated using the ConTEXTure model, being worn by its mesh. Each model is shown from three equidistant azimuth angles of 0\degree, 120\degree, and -120\degree, offset from the front. The prompts used to generate each texture are written directly beneath the images.}
    \label{fig:two_by_two_grid}
\end{figure*}

\section{Introduction}
Recent advancements in 3D modeling and texture synthesis \cite{richardson2023texture, cao2023texfusion, chen2023text2tex, metzer2023latent, liu2023text, ceylan2024matatlas, tang2024intex, wu2024texro, tang2023text, youwang2024paint, wang2023breathing, gao2024genesistex} have led to significant improvements in generating custom textures for any provided input mesh. TEXTure \cite{richardson2023texture}, a pioneering model in this domain, utilizes text prompts to generate texture atlases using the Stable Diffusion \cite{rombach2022high} depth pipeline, which leverages depth maps of the mesh rendered from diverse viewpoints. However, the images generated for different viewpoints often lack "consistency across all views."

Consistency across all views means that the images generated for each viewpoint (front, left, right, back, top, and bottom) is in harmony with each other. Each image accurately reflects the 3D mesh from its specific perspective without any discrepancies or distortions. This ensures that the texture map learned from the multiple images appears uniform and correct from every angle.

The problem that the generated multi-view images lack view-consistency may be caused by several factors. The main one seems that most images in the dataset used to train the depth pipeline are taken from the front viewpoint. This issue is referred to by \cite{liu2023zero} as the \emph{viewpoint bias} problem, demonstrating how both Dall-E 2 \cite{ramesh2022hierarchical} and Stable Diffusion, when provided the prompt ``\emph{a chair}'', consistently generate chair images in a forward-facing canonical pose. Even when augmenting the text prompts with directional information (e.g., ``\{front, back, left, right, overhead, bottom\} view''), the problem generally continues to persist.

\begin{figure}
\begin{subfigure}{0.11\textwidth}
  \centering
  \includegraphics[trim={2cm 3cm 2cm 0},clip,width=\linewidth,bb=0 0 724 724]{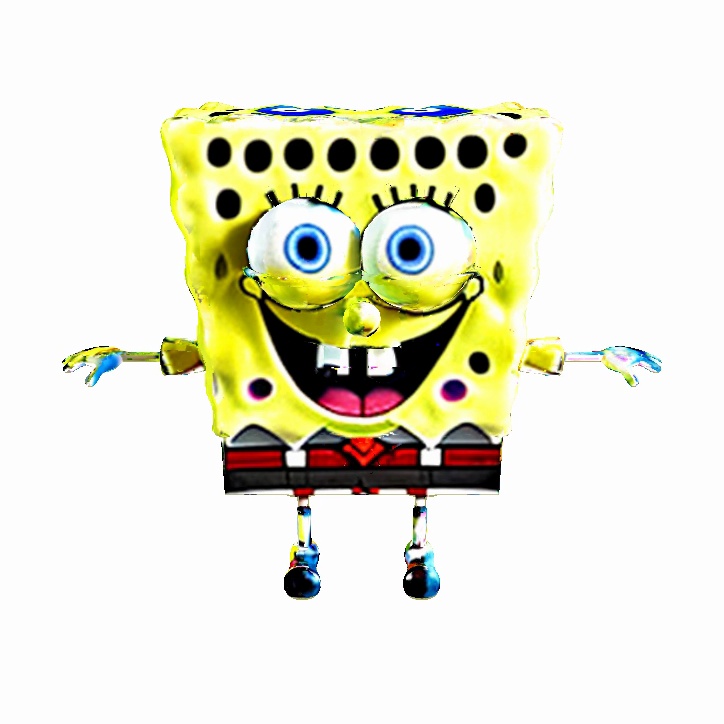}
  \caption{Front}
  \label{fig:spongebob_front}
\end{subfigure}%
\begin{subfigure}{0.11\textwidth}
  \centering
  \includegraphics[trim={2cm 3cm 2cm 0},clip,width=\linewidth,bb=0 0 724 724]{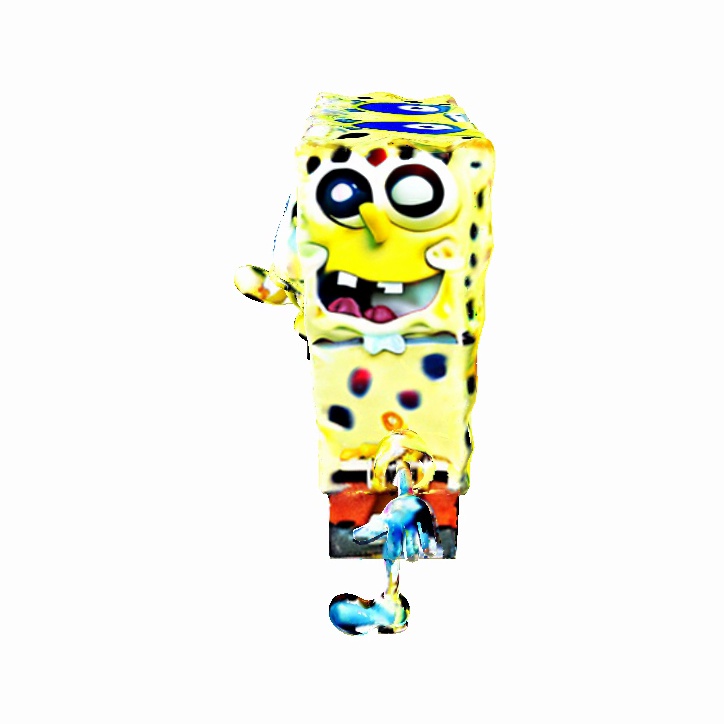}
  \caption{Left}
  \label{fig:spongebob_left}
\end{subfigure}
\begin{subfigure}{0.11\textwidth}
  \centering
  \includegraphics[trim={2cm 3cm 2cm 0},clip,width=\linewidth,bb=0 0 724 724]{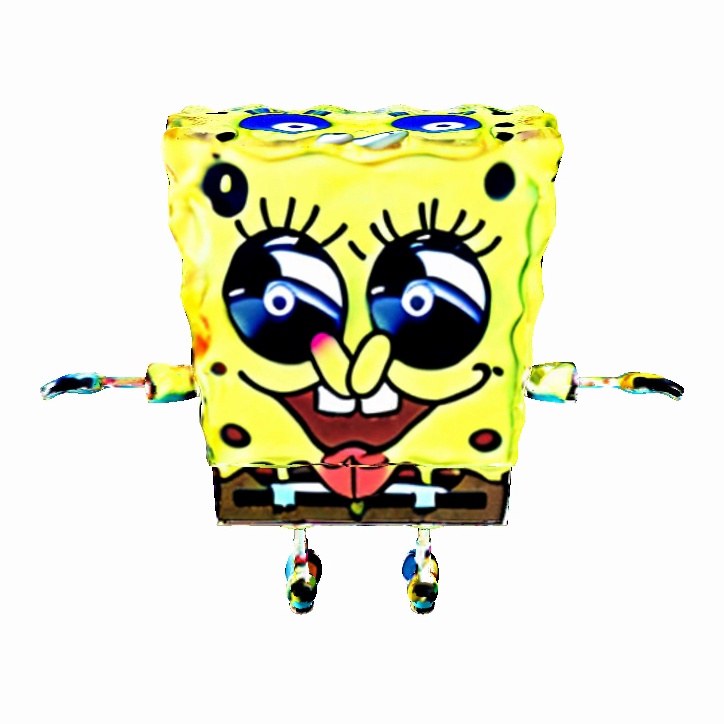}
  \caption{Back}
  \label{fig:spongebob_back}
\end{subfigure}
\begin{subfigure}{0.11\textwidth}
  \centering
  \includegraphics[trim={2cm 3cm 2cm 0},clip,width=\linewidth,bb=0 0 724 724]{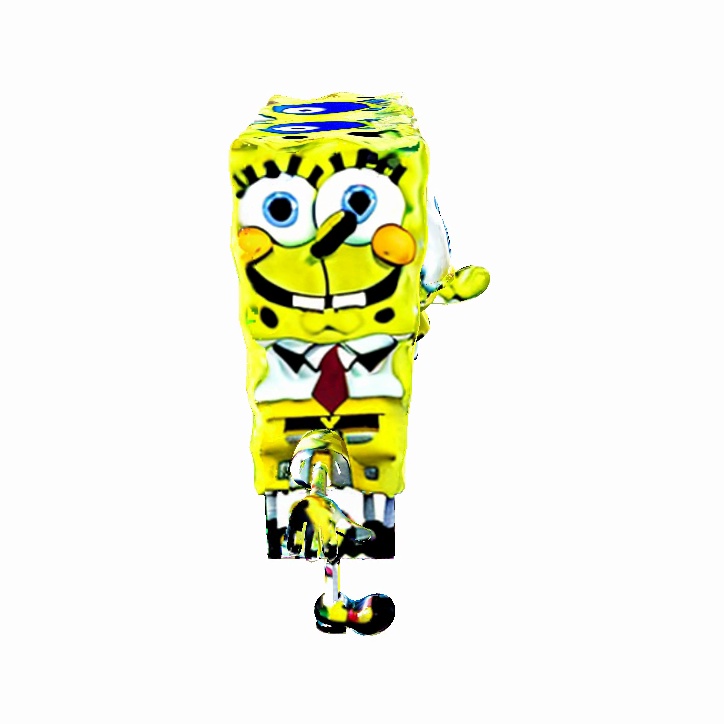}
  \caption{Right}
  \label{fig:spongebob_right}
\end{subfigure}

\caption*{Generated texture using TEXTure \cite{richardson2023texture}}

\begin{subfigure}{0.11\textwidth}
  \centering
  \includegraphics[trim={2cm 3cm 2cm 0},clip,width=\linewidth,bb=0 0 724 724]{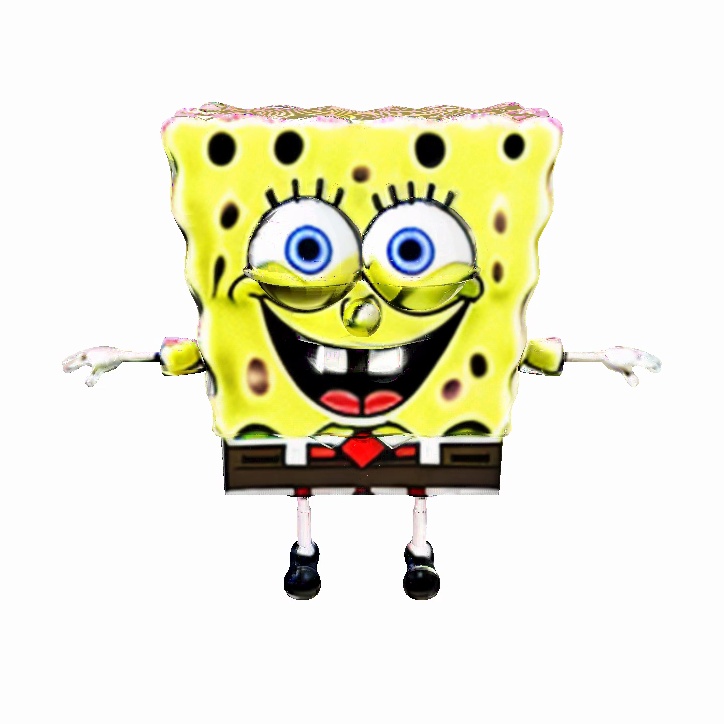}
  \caption{Front}
  \label{fig:spongebob2_front}
\end{subfigure}%
\begin{subfigure}{0.11\textwidth}
  \centering
  \includegraphics[trim={2cm 3cm 2cm 0},clip,width=\linewidth,bb=0 0 724 724]{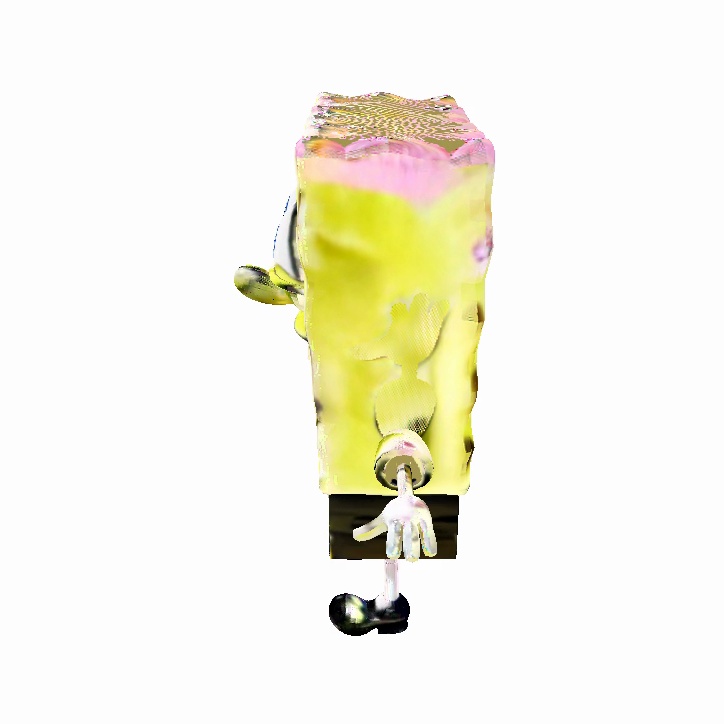}
  \caption{Left}
  \label{fig:spongebob2_left}
\end{subfigure}
\begin{subfigure}{0.11\textwidth}
  \centering
  \includegraphics[trim={2cm 3cm 2cm 0},clip,width=\linewidth,bb=0 0 724 724]{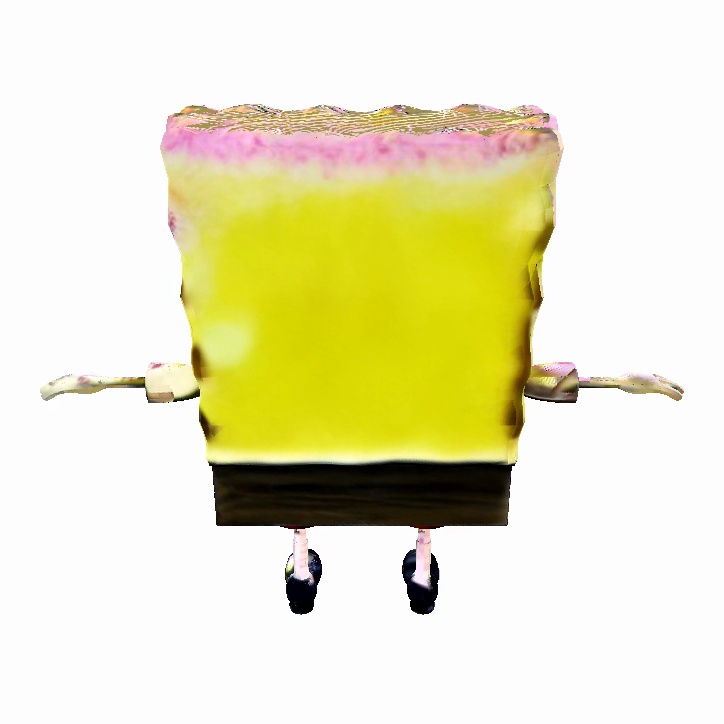}
  \caption{Back}
  \label{fig:spongebob2_back}
\end{subfigure}
\begin{subfigure}{0.11\textwidth}
  \centering
  \includegraphics[trim={2cm 3cm 2cm 0},clip,width=\linewidth,bb=0 0 724 724]{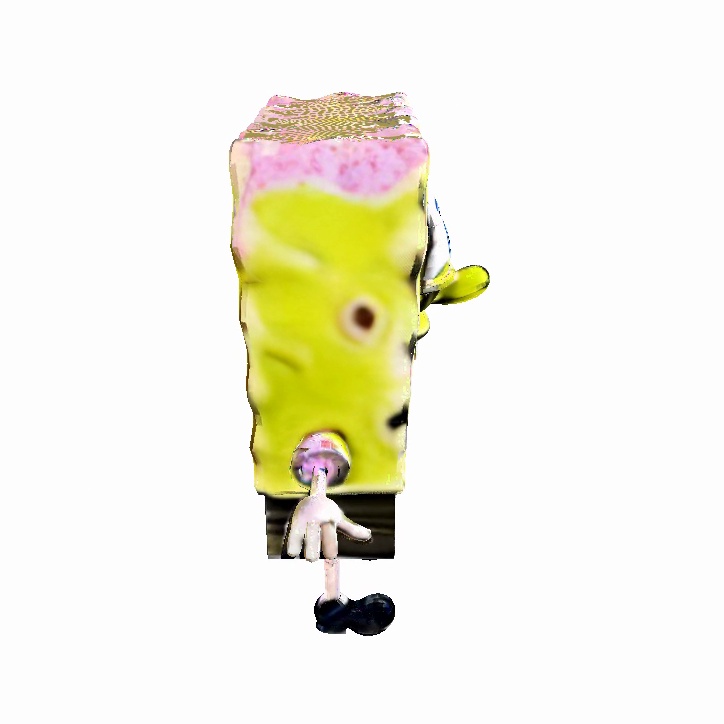}
  \caption{Right}
  \label{fig:spongebob2_right}
\end{subfigure}

\caption*{Generated texture using our model}

\caption{Texture generated using the original TEXTure model with the prompt ``A photo of Spongebob, \{\} view" and a guidance scale of 10. Note that the unique geometry of the SpongeBob character causes faces to be erroneously generated on each side of the mesh. This problem was resolved in our model.}
\label{fig:spongebob}
\end{figure}

\begin{figure*}
  \includegraphics[width=\textwidth,bb=0 0 1285 638]{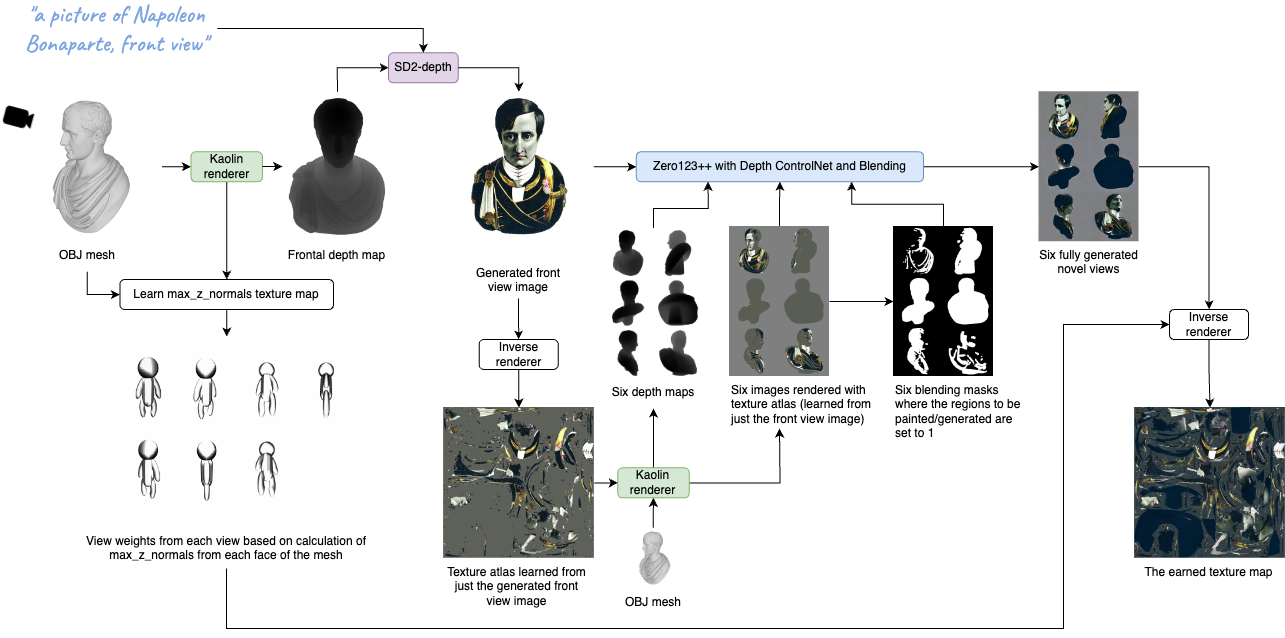}
  \caption{A visual representation of the ConTEXTure model's generation process. A latent image blending process is performed in our custom implementation of Zero123++ which is shown in more detail in Figure \ref{fig:zero123plus_blending}. Refer to Algorithm \ref{alg:contexture} for more information on the overall texture generation process.}
\end{figure*}

In our investigation of texture map generation, it was found that the problem of view-consistency is generally caused by two factors. First, the less detailed polygon representation of the mesh causes the view-consistency problem. For example,  the case of the mesh of a Napoleon Bonaparte bust, it has a high level of polygonal detail. It plays a role in facilitating the depth pipeline's performance by providing a rich set of cues for image generation from the text prompt. Conversely, cartoon-style meshes with fewer details present ambiguity in depth maps, particularly for identifying distinct features like the front or back of a character. Despite appending directional cues to the text prompt, the lack of detailed geometry can cause the generation of images which lack the consistency among views.

Second, the view-consistency problem stems from inherent viewpoint biases within the Stable Diffusion dataset. As seen in Figure \ref{fig:spongebob}, this is particularly evident in generating textures for characters like SpongeBob from non-standard viewpoints. SpongeBob is selected as an exemplar due to his distinct geometry, where his front and back sides become invisible when viewed from a perpendicular 90-degree angle. This unique characteristic accentuates the challenges in depth-based image generation, diverging significantly from typical human or object meshes, thereby providing a stringent test case for assessing the robustness of our texture synthesis approach in handling atypical shapes and viewpoints. Given the model's training on diverse internet-scale datasets, it has a predisposition towards generating recognizable features, such as faces, even when provided with depth maps from unconventional viewpoints. This bias leads to the erroneous placement of these features across multiple sides of a mesh, highlighting the limitations of relying solely on text prompts and standard depth cues in guiding the texture synthesis process for characters with iconic and highly recognizable features.

To address this problem, we introduce the ConTEXTure model, which adapts the TEXTure model to use the Stable Diffusion depth pipeline (hereinafter referred to as \texttt{SD2-depth}) alongside the version of Zero123++ \cite{shi2023zero123++} fine-tuned with ControlNet \cite{zhang2023adding} to support depth-controlled novel view image generation. Depth maps play a crucial role in texture atlas generation, guiding the texture synthesis to ensure alignment with the mesh's geometry. \texttt{SD2-depth} is used to generate just the front view of the object. Then we learn the texture atlas from the front view image, and renders the mesh using the partially learned texture atlas  from six  viewpoints, producing six viewpoint images. The front view image is reused as a condition image to generate all remaining viewpoints using the depth-controlled Zero123++. The six render images of the mesh are also used as a guide to the process of generating the six viewpoint images by the depth-controlled Zero123++.

Zero123++ improves upon its predecessor models Zero 1-to-3 \cite{liu2023zero} and Zero123-XL \cite{deitke2024objaverse} by simultaneously generating six diverse viewpoint images in one denoising loop as opposed to one viewpoint at a time, significantly speeding up the texture map generation process. This simultaneous generation of viewpoint images led us implement an inverse renderer that learns the texture atlas from the  multi-view images as the target images simultaneously. It is a departure from the approach used in \cite{richardson2023texture} which performs inverse rendering one viewpoint image at a time. This approach initializes the texture map image as  learnable parameters, adjusting them by reducing the loss between the given seven target viewpoint images (including the front viewpoint image) and the seven viewpoint images rendered from the textured mesh rendered with the Kaolin \cite{jatavallabhula2019kaolin} library. Several modifications had to be made to the loss equation to support simultaneous multi-view inverse rendering. This new modified approach is a natural one given the novel architecture of Zero123++.

\section{Related works}

\subsection{Diffusion models}
Although GANs historically held state-of-the-art performance for image generation tasks \cite{brock2018large, karras2020analyzing}, diffusion models have recently proven to be a strong, superior alternative \cite{dhariwal2021diffusion} and have since overtaken GANs in popularity. Widely used image generation models like Stable Diffusion \cite{rombach2022high} and DALL-E \cite{ramesh2022hierarchical, betker2023improving} are based on the diffusion model architecture. Stable Diffusion is popular due to it being open source and relatively easy to fine-tune with various types of conditions. Most applications of image generation can be classified as text-to-image, but other conditions can be used in conjunction with text prompts. Stable Diffusion additionally provides a depth-to-image pipeline. The depth map condition of the pipeline can be created from an image by means of MiDaS \cite{Ranftl2020, Ranftl2021} or rendered using a given mesh from a given viewpoint. 

Although Stable Diffusion can be fine-tuned directly, it is a known issue where fine-tuning the diffusion model with a small dataset can destroy its performance. This problem was resolved with the release of ControlNet \cite{zhang2023adding}. This allows for fine-tuning to take place in the copied backbone while the weights on the original model are preserved.

\subsection{Texture generation models}
The development of texture generation models has evolved significantly, with diffusion models leveraging large-scale data marking a notable leap forward. Early models like AUV-Net \cite{chen2022auv} and Texturify \cite{siddiqui2022texturify} facilitated 2D texture generation on 3D meshes by learning aligned UV spaces and training 3D StyleGANs. These models effectively addressed challenges in achieving global coherence and detailed texture representation.

PS-NeRF \cite{10.1007/978-3-031-19769-7_16} estimates the geometry, PBR maps, lights from multiple viewpoint images of a non-Lambertian object under multiple unknown directional lights. From the learned volume-based radiance field, PS-NeRF constructs the surface normals, spatially-varying BRDFs, and lights by means of a shadow-aware differentiable rendering layer. While sophisticated PS-NeRF requires multiple view-consistent images, which are not easy to acquire. We aim at a simpler method of learning a texture atlas from a given mesh and a text prompt.

Diffusion models have ushered in a new era in texture generation. DreamFusion \cite{poole2022dreamfusion} and Magic3D \cite{lin2023magic3d} leveraged Score Distillation Sampling (SDS) to optimize 3D representations using large-scale 2D text-to-image diffusion models, producing high-quality textures guided by text prompts. However, these models required lengthy optimization processes and faced limitations in color saturation and diversity due to high guidance weights \cite{ho2022classifier}.

The TEXTure model \cite{richardson2023texture} revolutionized texture synthesis by employing depth-conditioned diffusion models to iteratively paint 3D surfaces from different viewpoints, significantly enhancing efficiency and texture quality. This method enabled the rapid generation of high-quality textures without extensive optimization,offering substantial speed improvements.

Text2Tex \cite{Chen_2023_ICCV} resembles TEXTure in many ways, synthesizing partial textures from multiple viewpoint images. It introduces an automatic view sequence generation method to determine the next best view for updating the texture atlas.

TexFusion \cite{cao2023texfusion} proposes constructing the texture atlas in the latent space while multi-view images are being denoised. At each denoising step, each viewpoint image is sequentially denoised, and the noisy version of the texture atlas is learned from the currently denoised viewpoint images. This contrasts with previous approaches, which learn the texture atlas from a set of clean images corresponding to multiple viewpoints. It leads to more smoothed transition between the regions of the texture atlas corresponding to different view images.

However, these models continue to suffer from viewpoint bias due to the use of diffusion models trained on disproportionate amounts of images with front-facing directions. TexRO \cite{wu2024texro} is similar to the model created in our research in that it also utilizes Zero123++ \cite{shi2023zero123++}, a model used for novel view synthesis, to generate the texture for a given mesh. Although the viewpoint bias problem was not explicitly mentioned in the paper, its usage of Zero123++ would lead to the problem being resolved here. However, the inverse rendering in TexRO is not simultaneous and uses an effective but counter-intuitive recursive optimization approach to increase details in the resulting texture atlas.

\subsection{Novel view synthesis}
Here we review techniques of synthesis of multiple view images which are the backbone of the construction of texture atlas.  Recent advancements in novel view synthesis have significantly contributed to the field of 3D reconstruction and image generation. Among these, the development of the Zero 1-to-3 model \cite{liu2023zero} was specifically designed to address the "viewpoint bias" problem commonly encountered in text-to-image models, such as Stable Diffusion \cite{rombach2022high} and DALL-E 2 \cite{ramesh2022hierarchical, betker2023improving}, where generated images often display objects in canonical poses regardless of the specified viewpoint.

Prior to Zero 1-to-3, novel view synthesis primarily leveraged neural radiance fields (NeRFs) \cite{mildenhall10representing} for rendering specific scenes from novel viewpoints. NeRF-based methods, including DreamFields \cite{jain2022zero}, utilized a neural representation that required multiple views of an individual scene for accurate synthesis, limiting their application to pre-defined scenes. These methods demonstrated high fidelity in view synthesis but were constrained by the need for extensive 3D annotations and specific scene data.

The introduction of Zero 1-to-3 marked a significant shift by enabling zero-shot novel view synthesis and 3D shape reconstruction from a single RGB image. This model utilizes large diffusion models trained on extensive 2D image datasets, allowing it to generalize to unseen images and achieve state-of-the-art results without the need for 3D supervision. The model's ability to fine-tune on synthetic datasets and manipulate camera viewpoints through learned geometric priors has set a new benchmark in the field.

Building on the success of Zero 1-to-3, subsequent models like Zero123-XL \cite{deitke2024objaverse} and Zero123++ \cite{shi2023zero123++} have further enhanced the capabilities of novel view synthesis. Zero123-XL leverages the extensive Objaverse-XL dataset \cite{deitke2024objaverse}, comprising over 10 million 3D objects, to improve generalization and performance across a wide range of modalities, including photorealistic assets, cartoons, and sketches. Zero123++ integrates depth control through a Depth ControlNet \cite{zhang2023adding}, enhancing its ability to handle complex geometric variations and achieve superior image quality metrics such as LPIPS \cite{zhang2018unreasonable}. This model not only improves on the synthesis quality but also extends the application to more diverse and challenging datasets, showcasing the robustness and adaptability of modern novel view synthesis techniques.

\begin{figure*}[h]
  \includegraphics[width=0.8\textwidth]{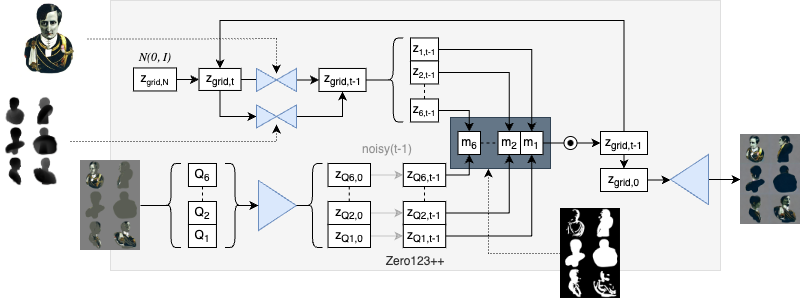}
  \caption{Our usage of Zero123++ features a custom diffusion process that was inspired by the implementation of \cite{richardson2023texture}. We perform blending of the latent image $z_{grid,t-1}$ with noisy ground truth latent $z_{Q_{grid,t-1}}$ after each denoising iteration $t-1$. Performing blending according to the blending mask $m_{grid}$ ensures that the front view image that was already projected back unto the texture atlas is not regenerated by the Zero123++ model.}
  \label{fig:zero123plus_blending}
\end{figure*}

\section{Method}

\begin{algorithm}
\caption{ConTEXTure Algorithm}
\label{alg:contexture}
\begin{algorithmic}[1]
    \State \textbf{Input:} Text prompt $text$, Mesh $M$, camera viewpoints $\{v_0, \ldots, v_6\}$ where $v_0$ is the front viewpoint
    \State Learn the $max\_z\_normals$ meta-texture map \label{lst:line:maxnormals}
    \State $D_0$ $\gets$ Render depth map of $M$ at $v_0$
    \State $z_{0, N_{sd2}} \gets \mathcal{N}(0, I)$
     \For{$t = N_{sd2}$ \textbf{to} $0$}
     \State $z \gets \mathcal{N}(0, I)$
      \State $z_{0, t-1} \gets \text{TEXTure\_denoise} (z_{0,t},text, D_{0}, t, z)$
      \EndFor
   
    \State $Q_{0} \gets x_{0, 0} \gets \text{SD2\_VAE\_decode}(z_{0, 0})$

    \State $T \gets$ Project-back $Q_0$ to learn $T$ for $Q_0$ \label{lst:line:learntexture1}
    \For {$i = 1 \ldots 6$}
        \State $D_i \gets \text{Render depth map of $M$ at $v_i$}$
        \State $m_i \gets $ Render object mask of $M$ at $v_i$ where the region of $M$ projected by the learned part of $T$ is set to 0
        \State $Q_i \gets \text{Render image of $M$ wearing $T$ at $v_i$}$
    \EndFor

    \State $D_{grid} \gets \text{Create 2x3 grid image from $D_1 \ldots D_6$}$
    \State $m_{grid} \gets \text{Create 2x3 grid image from $m_1 \ldots m_6$}$
    \State $Q_{grid} \gets \text{Create 2x3 grid image from $Q_1 \ldots Q_6$}$
    \State $z_{Q_{grid},0} \gets \text{Zero123plus\_VAE\_encode($Q_{grid}$)}$

    \State $ z_{grid, N_{zero123}} \gets \mathcal{N}(0,I)$
    \For {$t = N_{zero123}$ ... $0$}
        \State $z_{grid, t-1} \gets \text{Zero123plus\_denoise}(z_{grid,t}, Q_0, D_{grid}, t-1)$
        \State $z_{Q_{grid},t-1} \gets \text{Add the noise of $t-1$ to $z_{Q_{grid},0}$}$
        \State $z_{grid,t-1} \gets z_{grid,t-1} \cdot m_{grid} + z_{Q_{grid},t-1} \cdot (1 - m_{grid})$ 
    \EndFor

    \State $x_{grid, 0} \gets \text{Zero123plus\_VAE\_decode}(z_{grid, 0})$
    \State $x_{1, 0}, x_{2, 0}, x_{3, 0}, x_{4, 0}, x_{5, 0}, x_{6, 0} \gets \text{Split $x_{grid, 0}$}$
    
    \State $T \gets$ Project-back all view images $x_{i,0} ,i = 0 \ldots 6,$ to learn the texture map thereof, using $max\_z\_normals$ as the weights of each view image to contribute to the texture map \label{lst:line:learntexture2}
\end{algorithmic}
\end{algorithm}

\subsection{Depth-based diverse view image generation}

\begin{figure}[t]
    \begin{subfigure}{0.23\textwidth}
  \centering
  \includegraphics[trim={1cm 4.5cm 1cm 0},width=\linewidth,bb=0 0 724 724]{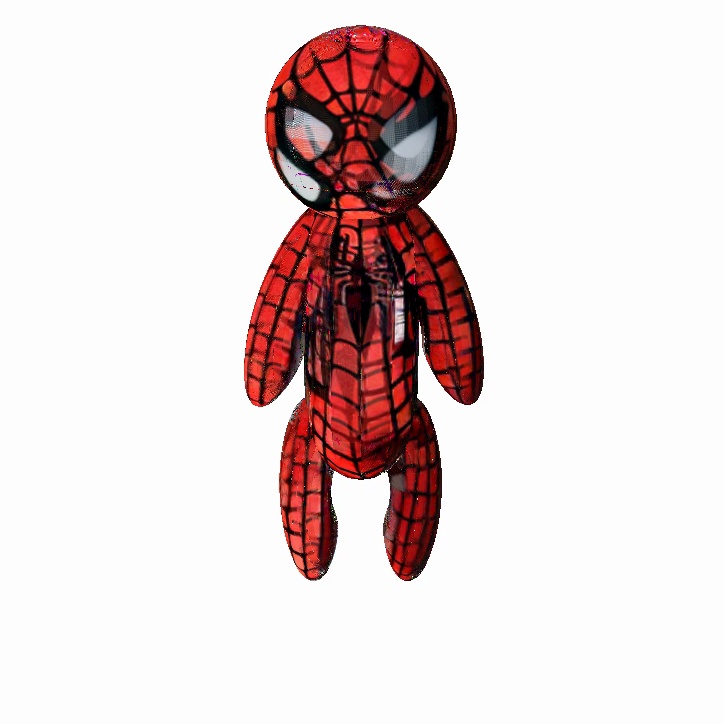}
  \caption{Without blending with the ground truth render image}
  \label{fig:unblended}
\end{subfigure}%
\begin{subfigure}{0.23\textwidth}
  \centering
  \includegraphics[trim={1cm 4.5cm 1cm 0},width=\linewidth,bb=0 0 724 724]{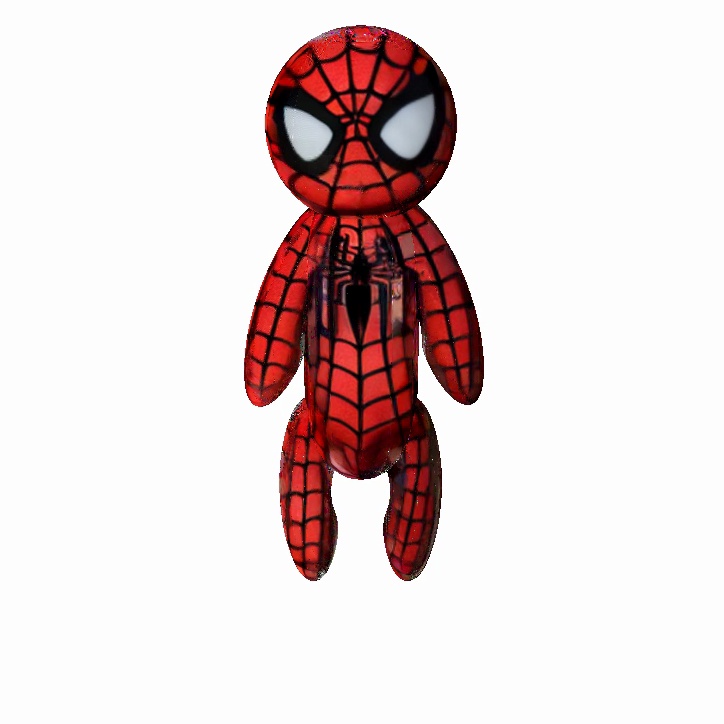}
  \caption{With blending with the ground truth render image}
  \label{fig:blended}
\end{subfigure}
\caption{There are slight misalignments among the novel view images by Zero123++ which result in a low quality texture when projected onto the mesh without further postprocessing. The blending technique, used in \cite{richardson2023texture} for preventing already-generated regions of the image from being overwritten, continues to prove effective in ConTEXTure. Prompt of ``A photo of Spiderman, front view" was used for both textures.}
\label{fig:blend_compare}
\end{figure}

The original TEXTure model is designed to generate diverse view images sequentially, using \texttt{SD2-depth}. This model generates a novel view image from the depth map of the mesh, given a specific viewpoint and a text prompt that includes the relevant view descriptor (e.g., "Spiderman, left view"). Initially, TEXTure "projects back" the front view image to the texture atlas. This process involves learning the texture atlas as trainable parameters of the neural network from the front view image by minimizing the loss between the target front view image and the image rendered using the current texture atlas.

Subsequently, the next view image is generated using the depth pipeline. The texture atlas is further refined by minimizing the loss between this new view image and the image rendered from the updated texture atlas. At this stage, the region of the texture atlas is re-learned only if the region of the mesh onto which it is mapped is better viewed from the current viewpoint compared to the previous one. This condition functions as a view-weight over the mesh region, which may be observed from multiple viewpoints. The better a mesh region is viewed, the more significantly it contributes to the texture atlas.

TEXTure further refines the latent image for a given viewpoint, obtained through the depth pipeline, by incorporating the ground truth image of the mesh rendered from that specific viewpoint. Specifically, the latent image being denoised is blended with the ground truth rendered image as the background, similar to the in-painting pipeline where the image being denoised is blended with a background image. This approach ensures consistency between the generated and rendered images by preventing the denoising process from overwriting the valid parts of the texture atlas already learned from the prior viewpoint. The blending process utilizes the \emph{keep}, \emph{refine}, and \emph{generate} masks, which are created by comparing the z-normals (the z-coordinates of the faces projected onto the screen in the current camera) with a cache of the z-normals that holds the maximum z-normals from the previously considered viewpoints.

Since Zero123++ requires a condition image to generate six viewpoints, we decided to keep using \texttt{SD2-depth} to generate an image from the first viewpoint $v_0$, which we always assume to be the front view. This generated frontal image $Q_0$ is then used to create six novel views of the same object using Zero123++. We employed the \emph{Depth ControlNet for Zero123++} to control the generated image $Q_0$ with six depth maps $D_1 \dots D_6$) corresponding to the mesh from the six viewpoints $v_1 \dots v_6$. The text prompt $T_{front}$ used for \texttt{SD2-depth} has directional information (e.g.. ``A picture of Napoleon Bonaparte\textbf{, front view}'') concatenated to the end of the original prompt $T$. Zero123++ uses FlexDiffuse \cite{speed2022flexdiffuse} for utilizing global image conditioning in order to minimize the amount of fine-tuning required for Zero123++ to obtain the desired results. Although we tried using $T$ as the prompt when using Zero123++ in an attempt to improve prompt fidelity, there was a noticeable decrease in quality. We assume that this can be attributed to the fact that public versions of Zero123++ were not trained with the text prompt when using FlexDiffuse.

In ConTEXTure, the process of learning the texture atlas occurs in two steps. First, the texture atlas is partially learned from the front view image, and the ground truth render images are produced using this partially learned texture atlas. Second, the remaining part of the texture atlas is learned using the seven viewpoint images, including the front view image. Confer lines \ref{lst:line:learntexture1} and \ref{lst:line:learntexture2} of Algorithm \ref{alg:contexture}. From the perspective of the texture itself, learning the texture atlas from just $Q_0$ is redundant due to the subsequent learning of the entire texture atlas using $Q_0 \dots Q_6$. However, we found that Zero123++ has the tendency to produce view images that suffer from misalignment problems, leading to a low quality texture seen in Figure \ref{fig:unblended}. To resolve this problem, we borrowed the blending technique from TEXTure \cite{richardson2023texture} to prevent modification of the area corresponding to the front of the mesh. The texture map learned from $Q_0$ is used as the image to blend with the image being denoised by the Zero123++ pipeline. This blending process, which can be seen in Figure \ref{fig:zero123plus_blending}, forces the pipeline to leave the frontal region unchanged.

\begin{figure}[htbp]
    \captionsetup{skip=1pt}
    \centering
    \begin{minipage}{0.23\columnwidth}
        \centering
        \includegraphics[width=\textwidth,bb=0 0 256 256]{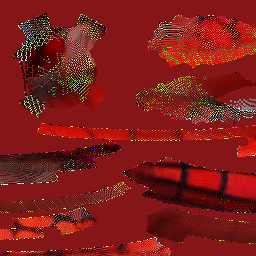}
    \end{minipage}
    \begin{minipage}{0.23\columnwidth}
        \centering
        \includegraphics[width=\textwidth,bb=0 0 256 256]{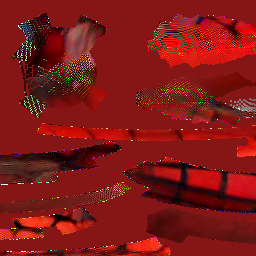}
    \end{minipage}
    \begin{minipage}{0.23\columnwidth}
        \centering
        \includegraphics[width=\textwidth,bb=0 0 256 256]{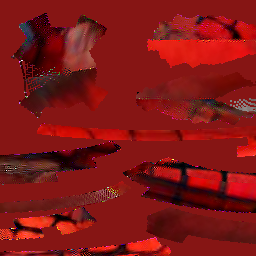}
    \end{minipage}
    \begin{minipage}{0.23\columnwidth}
        \centering
        \includegraphics[width=\textwidth,bb=0 0 256 256]{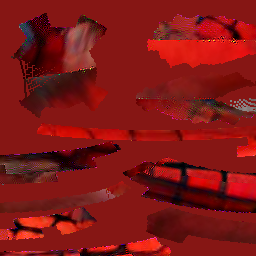}
    \end{minipage}
    \\
    \begin{minipage}{0.23\columnwidth}
        \centering
        \includegraphics[width=\textwidth,bb=0 0 64 64]{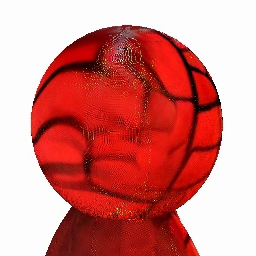}
        \caption*{$\alpha = 1$}
    \end{minipage}
    \begin{minipage}{0.23\columnwidth}
        \centering
        \includegraphics[width=\textwidth,bb=0 0 64 64]{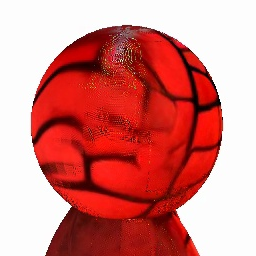}
        \caption*{$\alpha = 10$}
    \end{minipage}
    \begin{minipage}{0.23\columnwidth}
        \centering
        \includegraphics[width=\textwidth,bb=0 0 64 64]{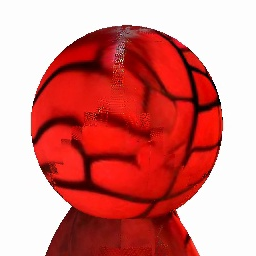}
        \caption*{$\alpha = 100$}
    \end{minipage}
    \begin{minipage}{0.23\columnwidth}
        \centering
        \includegraphics[width=\textwidth,bb=0 0 64 64]{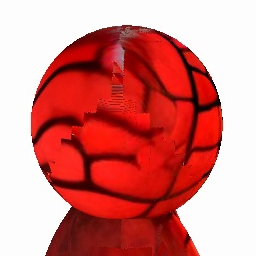}
        \caption*{$\alpha = \infty$}
    \end{minipage}
    \caption{A comparison between different values of $\alpha$. The first row shows a limited region of the generated texture map. The second row shows a close-up shot of the model when looked from the back. The version with $\alpha = \infty$ corresponds to the version using the binary mask. The mesh and texture shown here is the same as the one shown in Figure \ref{fig:blended}.}
    \label{fig:alpha-compare}
\end{figure}

\subsection{Simultaneous inverse rendering from multiple view images}

TEXTure and TexFusion are designed to generate images for multiple viewpoints one at a time. In contrast, the unique aspect of Zero123++ lies in its simultaneous generation of view images. This highlights the need for a new and more efficient approach for simultaneous view inverse rendering. However, performing inverse rendering without considering overlapping regions results in unnatural overlaps of inconsistent images in these areas. An obvious solution involves creating a binary mask for each viewpoint, which holds the maximum z-normals of the faces among all viewpoints. If the z-normals of a face in a given viewpoint are the same as the maximum z-normals for that face, then the face (and the corresponding pixel locations) is assigned a value of 1; otherwise, it is assigned 0. Confer Appendix \ref{binary_mask} for more information on the algorithm used to create these binary masks.

However, these binary view masks lead to the problem of seams in the overlapping regions of the texture atlas, requiring the edges to be smoothed to hide the resulting artifacts. To smooth these masks, we learn the meta-texture $N_i$, which holds the maximum z-normals among all viewpoints (line \ref{lst:line:maxnormals} of Algorithm \ref{alg:contexture}). When the values of $n_i$ (the z-normals at face $i$) are close to $N_i$, it indicates that these regions are better viewed by the camera and that their images are more valuable for contributing to the texture map. The meta-texture map $N_i$ is learned so that when projected onto the screen from each viewpoint, the projected $N_i$ is never less than $n_i$. The learning of $N_i$ using just the mesh and camera information, allowing it to be pre-computed before the generation of multi-view images and the learning of the texture atlas. Equation \ref{eq:max_z_normals_loss} is the loss used to train $N_{i_p}$, for every pixel $p$ among all viewpoints $i$.

\begin{equation} \label{eq:max_z_normals_loss}
\mathcal{L} = \sum_{i} \sum_{p} \text{ReLU} \left( n_{i, p} - N_{i, p} \right)
\end{equation}

After training of $N_i$, it is used to calculate view weights $W_i$ to be used during the learning of the texture map. The calculation of $W_i$ can be seen in Equation \ref{eq:view_weights}.

\begin{equation} \label{eq:view_weights}
W_i = \exp(-\alpha * |N_i - n_i|)
\end{equation}

$\alpha$ is a hyperparameter that determines the roughness of the amount of smoothing to be applied to the texture map. As can be seen in Figure \ref{fig:alpha-compare}, the resulting texture is smooth when $\alpha$ is set to be a low number. However, there is a problem where neon-colored pixels appear within the regions that have been projected back onto from more than one view image, and this issue is exacerbated for small values of $\alpha$. Multiple experiments led us to the conclusion that setting this value to around 10 leads to a fair balance between maximizing texture smoothness and minimizing the problem of neon-colored pixels. As $\alpha$ reaches $\infty$, the more similar the result should be to the binary masks.

\section{Experiments}

\subsection{Qualitative results}

We have not once been able to reproduce the viewpoint bias problem in constructing the texture atlas when using ConTEXTure. However, there remains several remaining issues. The first issue is the fact that some pixels of the texture map result in minor color distortions. This problem arises when a pixel on the texture map is contributed by the pixels of more than one view images. In a sense the multiple contributions to the texture map from multiple view images fight against each other, leading to unsatisfactory compromise. It means that the use of view-weights based on the difference between the z-normals of each viewpoint and the max z-normals meta-texture map is not completely satisfactory. A more sophisticated view-weights are needed. Note that this problem disappears completely when we use the binary view-masks to blend the contributions to the texture atlas from multiple view images. But it introduces the problem of seams between regions of the texture atlas. 

Another problem we encountered is that lighting information, such as shadows and lights, is \emph{baked into} the generated image. The reason for this is because Zero123++ was trained using rendered images of Objaverse models in an environment with random HDRI lighting \cite{shi2023zero123++}. This issue can be exacerbated when the generated front view image $Q_0$ has more bright or shadowy features than usual.

\subsubsection{Runtime}
In our environment, TEXTure takes 2 minutes and 54 seconds from start to finish, compared to 27 seconds when running ConTEXTure. Our model also surpasses TexFusion in terms of speed, which was reported to have a runtime of 2.2 minutes (approx. 2 minutes and 12 seconds) The biggest reason behind the speed difference is due to the fact that Zero123++ is used to generate six novel views in a single denoising process. This allows us to run just two denoising processes (once with \texttt{SD2-depth} and once with Zero123++), compared to ten \texttt{SD2-depth} denoising processes in TEXTure and eight in TexFusion. Although we are not able to run TexFusion in our runtime speed experiment due to the model being closed source, we used the same NVIDIA RTX A6000 GPU used by the author and also experienced an equivalent runtime speed when running TEXTure on our machine.

\subsection{Quantitative results} \label{quantitative-results}

\subsubsection{FID score}

\begin{figure}
    \begin{subfigure}{0.2\textwidth}
  \centering
  \includegraphics[trim={6.5cm 4.5cm 6.5cm 0},width=\linewidth,bb=0 0 1200 1200]{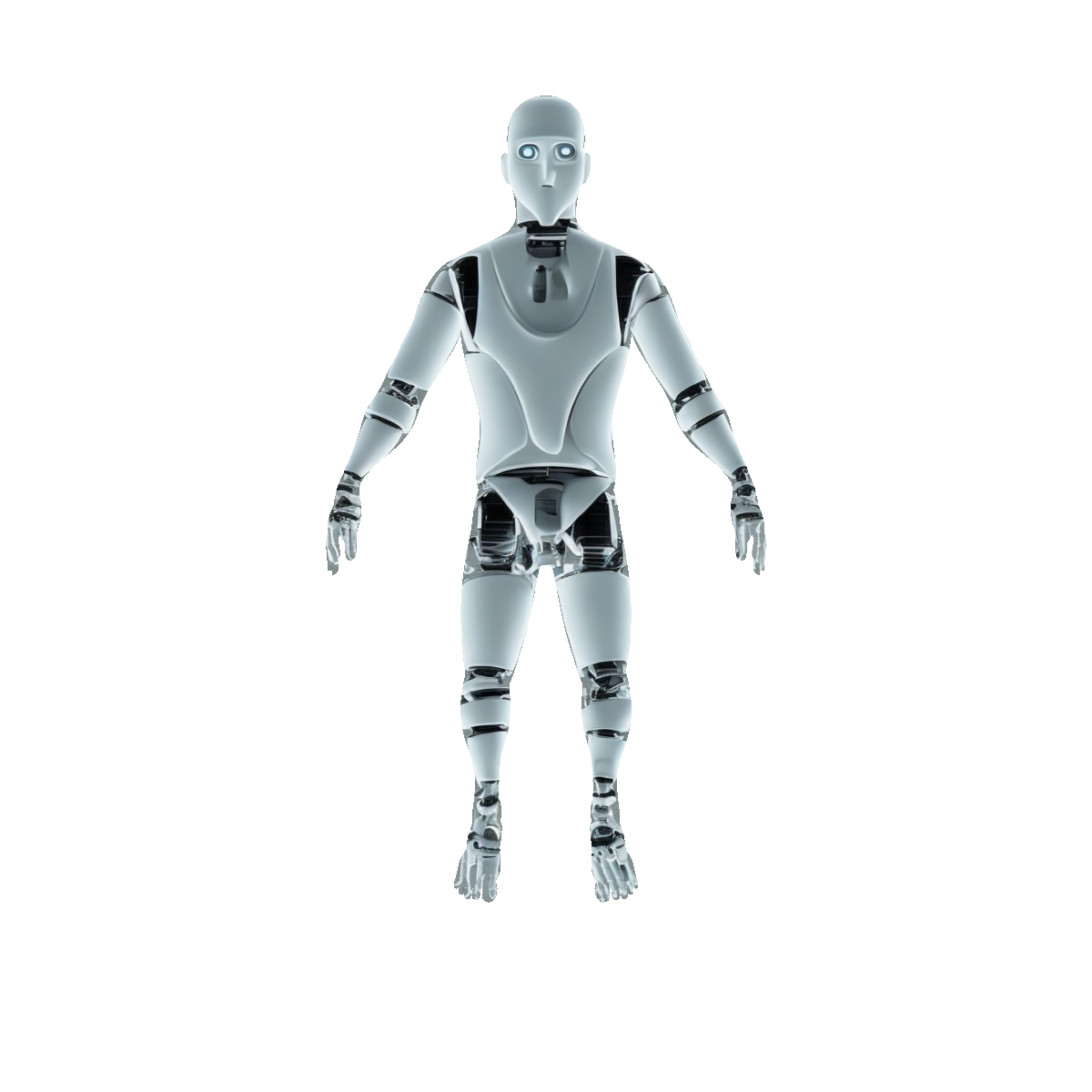}
  \caption{Front}
  \label{fig:proxy_front}
\end{subfigure}%
\begin{subfigure}{0.2\textwidth}
  \centering
  \includegraphics[trim={6.5cm 4.5cm 6.5cm 0},width=\linewidth,bb=0 0 1200 1200]{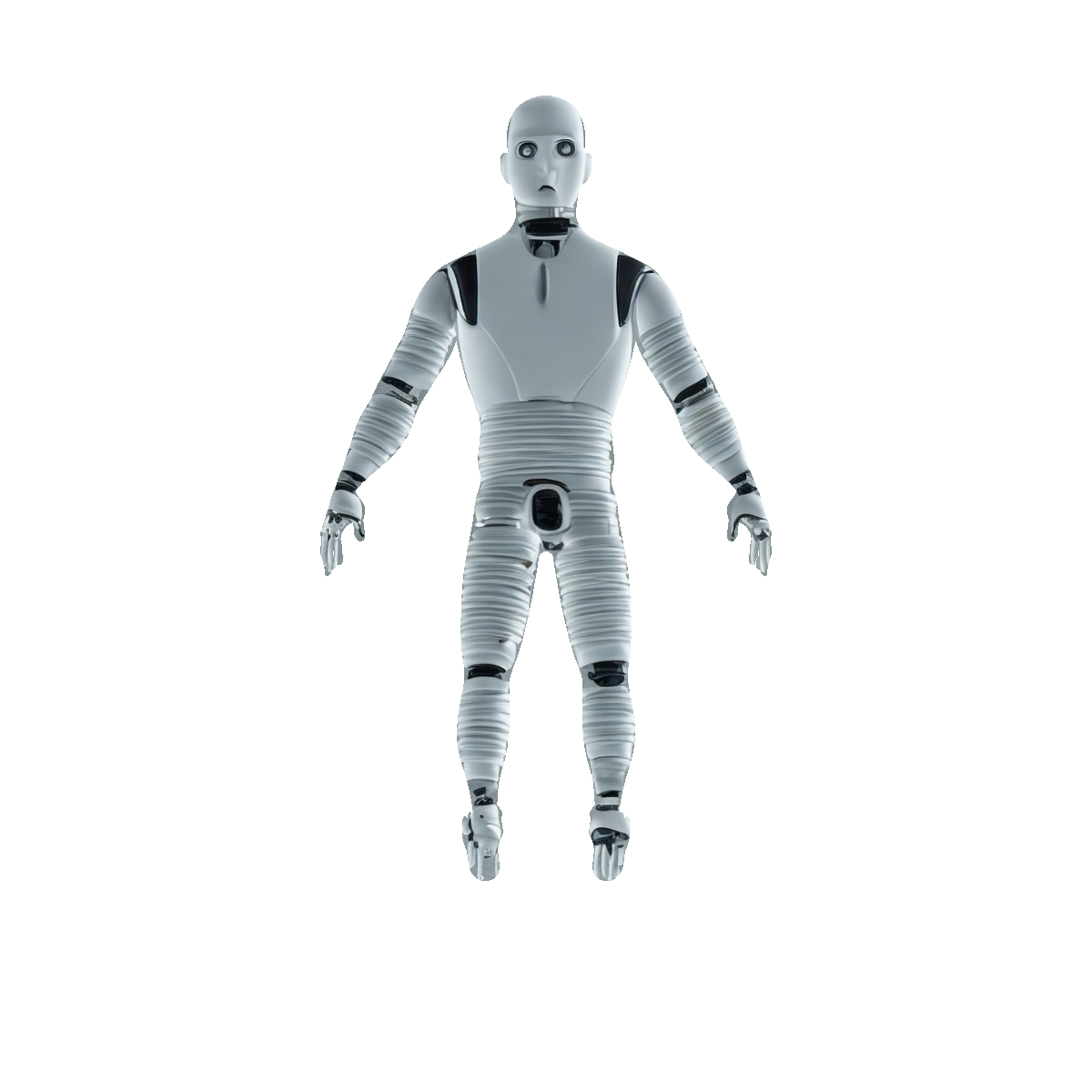}
  \caption{Back}
  \label{fig:proxy_back}
\end{subfigure}
\caption{When using the front- and back-side depth maps on the \emph{person} mesh using the prompt \emph{``white humanoid robot, movie poster,
villain character of a science fiction movie,"} the viewpoint bias issue manifests in the generation of eyes on both the front and back side of the head.}
\label{fig:proxy_set_bias}
\end{figure}

\begin{table}[ht]
\centering
\begin{tabular}{@{}llcc@{}}
\toprule
\multirow{2}{*}{Criteria} & \multirow{2}{*}{} & \multicolumn{2}{c}{Model} \\ \cmidrule{3-4}
                          &                   & TEXTure & ConTEXTure \\ \midrule
\multirow{3}{*}{User study} & Overall quality    & \textbf{57.5\%}     & 42.5\%       \\
                          & Viewpoint consistency & 47.5\%     & \textbf{52.5\%}       \\
                          & Prompt fidelity     & \textbf{50.0\%}     & \textbf{50.0\%}       \\ \midrule
FID                       &                     & 108.5    & \textbf{97.38}       \\ \midrule
Runtime                   &                     & 2 min 54 sec   & \textbf{0 min 27 sec}      \\ \bottomrule
\end{tabular}
\caption{Comparison of TEXTure and ConTEXTure models. The user study was conducted with 30 participants who were asked to compare 40 pairs of results from both models.}
\label{tab:comparison}
\end{table}

It is an inherently difficult task to evaluate the quality of a mesh-wearable texture map, which is perhaps why the original TEXTure model does not include quantitative results other than a user survey \cite{richardson2023texture}. The TexFusion model calculated the Fr\'echet Inception Distance (FID) \cite{heusel2017gans} to compare eight images of a proxy ground truth set with eight rendered images of the texture when worn by the mesh \cite{cao2023texfusion}. This proxy ground set consists of eight images generated by using eight corresponding depth maps of the mesh with \texttt{SD2-depth}. However, images generated in this manner also suffer from viewpoint bias (see Figure \ref{fig:proxy_set_bias}), demonstrating that using the same evaluation technique is not suitable in measuring how well we resolved the viewpoint bias problem. To circumvent this flaw in the proxy set, we used \texttt{SD2-depth} to generate an image from just the frontal viewpoint. We used this image with six depth maps extracted from non-frontal viewpoints as the condition image for Zero123++. The generated six view images were grouped together with the initial front view image and saved as seven separate image files. We then compared these images with the rendered images of the mesh wearing the generated texture map from the same viewpoints. TexFusion reported high FID scores and that the cause of this is due to the small size of the proxy ground truth set \cite{cao2023texfusion}. Our calculated FID scores shown in Table \ref{tab:comparison} are also considerably high as the numbers reported in TexFusion.

\subsubsection{User study} \label{userstudy}

Along with calculation of the FID score, we conducted a survey on Google Forms in which respondents are to look at and compare images for two textured meshes per section.  One texture is generated with \cite{richardson2023texture} and the other from ConTEXTure. To reduce bias in the answers provided by respondents, the order in which the two textures are shown is shuffled in each part. Each textured mesh is shown with four images and shows the front, left, back, and right side. For each textured mesh pair, 30 respondents were asked to choose between the two textures using the following criteria:

\begin{enumerate}
  \item \textbf{Overall quality} - The higher the quality, the more natural the colors and details appear, and the fewer unnatural artifacts there are.
  \item \textbf{Viewpoint consistency} - The consistency is higher if features such as the eyes, nose, and mouth are not positioned on the back or sides of the mesh.
  \item \textbf{Prompt fidelity} - The higher the fidelity, the more accurately the texture image reflects the prompt.
\end{enumerate}

A total of 30 respondents were provided with an explanation on the meaning of \textit{viewpoint consistency} and \textit{prompt fidelity}. The survey included 40 texture pairs, where each pair included the generated texture results from both TEXTure and ConTEXTure, the prompt used to generate the two textures, and a question pertaining to each of the three criteria. Every question required the respondent to choose which texture is superior in regards to the criterion. Refer to Appendix \ref{userstudydetails} for more details.

The results of the user study demonstrated that ConTEXTure outperforms TEXTure in viewpoint consistency, which demonstrates its superiority in addressing the viewpoint bias problem. However, it was tied in terms of prompt fidelity and fell short in regards to the overall quality. We analyzed the reasons this may have been the case and realized that \texttt{SD2-depth} generates images with more vibrant colors and more realistic details in object textures. Generated textures using TEXTure that do not visibly suffer from viewpoint bias had the tendency to score higher in the survey.

\section{Conclusion}

In this paper, we presented a solution to the viewpoint bias problem that was originally solved in novel view synthesis by Zero123 \cite{liu2023zero} but had yet to be attempted in texture atlas generation. Throughout the course of this research, we discovered that the viewpoint bias problem is pervasive throughout this domain, requiring us to make modifications to the evaluation process presented in \cite{cao2023texfusion}.

\subsection{Future work}

Although we were able to use Zero123++ for texture atlas generation, its tendency to generate texture faces with shadows prevents it from being seen as a pure texture generator. Although this does not end up being a problem when the objective is to reconstruct a novel view of an object in the condition image, this leads to problems in texture generation. Any future research would have to involve separating lighting information from the base color information of a given texture map. This will allow the image generation model to have control over the lighting information.

The issue with undesirable artifacts on the generated texture atlas frequently seen in TEXTure continues to be noticeable in ConTEXTure. Improving texture generation to be more high quality and globally coherent was well achieved in TexFusion. Although the code for TexFusion is closed source and our approach largely diverges from that of TexFusion primarily due to our usage of simultaneous multi-view inverse rendering, several approaches such as interleaving and usage of latent texture maps would very likely improve ConTEXTure in a similar manner.

As expressed in Section \ref{quantitative-results}, using a ``proxy ground truth dataset" for evaluation, an approach that was originally proposed in \cite{cao2023texfusion}, has its shortcomings. This demonstrates a need for a more neutral evaluation method, which we may attempt to create for future research.

%
%
%
%

\bibliographystyle{ACM-Reference-Format}
\bibliography{sample-bibliography}

\appendix

\section{Z-normal binary mask generation} \label{binary_mask}

Since ConTEXTure uses multiple novel view images to learn the texture map in a simultaneous fashion, it is crucial to determine which images should apply to which areas of the mesh. We follow the basic idea that every face of the mesh should be \emph{assigned} to it the viewpoint in which its z-normal value is maximized. To create this binary face view mask, we used a vectorized version of Algorithm \ref{alg:binary_mask}.

\begin{algorithm}
\caption{Binary mask creation}
\label{alg:binary_mask}
\begin{algorithmic}[1]
    \State \textbf{Input:} Face index $F$, normal map $N$
    \State $M \gets \text{\{\}}$
    \State $(V, H, W) \gets \text{shape of } F$
    \State $mask \gets \text{Boolean matrix of shape } (V, H, W)$
    
    \For {$v = 0 \text{ to } V-1$}
        \For {$i = 0 \text{ to } H-1$}
            \For {$j = 0 \text{ to } W-1$}
                \State $f \gets F[v, i, j]$
                \If {$f \text{ is valid}$}
                    \If {$f \notin M$}
                        \State $M[f] \gets \text{\{\}}$
                    \EndIf
                    \If {$v \notin M[f]$}
                        \State $M[f][v] \gets \text{[]}$
                    \EndIf
                    \State Append $(i, j)$ to $M[f][v]$
                \EndIf
            \EndFor
        \EndFor
    \EndFor

    \For {$f \text{ in } M$}
        \If {$M[f] \text{ is not empty}$}
            \State $V_f \gets \text{keys of } M[f]$
            \State $Z_f \gets N[V_f, 2, f]$
            \State $z_{\max} \gets \max(Z_f)$

            \For {$v \text{ in } M[f]$}
                \If {$N[v, 2, f] < z_{\max}$}
                    \State $i, j \gets M[f][v]$
                    \State $mask[v, i, j] \gets \text{False}$
                \EndIf
            \EndFor
        \EndIf
    \EndFor

\State \textbf{Output:} $mask$
\end{algorithmic}
\end{algorithm}

\section{User study details} \label{userstudydetails}
The texture prompts used in the survey are listed in Table \ref{table:prompts}. Since each texture pair includes three questions, there is a total of 120 questions in the full survey. Due to the sheer length of the full survey, we created one unique survey for each respondent where each survey asks about a random selection of 10 prompts. After 30 respondents completed the survey, we calculated the ratings for each texture. For each criteria of each texture pair, the texture with more votes was given one point. Thus, the denominator of the percentages in the user study results of Table \ref{tab:comparison} is the number of prompts in the survey, or 40. Because most of the survey respondents were Korean native speakers, all English text prompts were translated to Korean using ChatGPT 4o.

\begin{table*}
\centering
\begin{tabular}{|l|l|}
  \hline
  \textbf{Mesh} & \textbf{Text Prompts} \\
  \hline
  \begin{tabular}[t]{@{}l@{}}
    \texttt{person.obj} \\
    (Text2Mesh \cite{michel2022text2mesh}) \\
  \end{tabular} & \begin{tabular}[t]{@{}l@{}}
    \textbf{comic book superhero, red body suit} \\
    white humanoid robot, movie poster, villain character of a science fiction movie \\
    futuristic soldier, glowing armor, protagonist of an action game \\
    steampunk adventurer, leather attire with brass accessories \\
    astronaut in a sleek space suit, exploring alien worlds \\
    cyberpunk hacker, neon-lit clothing, main character in a dystopian cityscape \\
    \end{tabular} \\
  \hline
  \begin{tabular}[t]{@{}l@{}}
    \texttt{rp\_alvin\_rigged\_003\_yup\_a.obj} \\
    (Renderpeople) \\
  \end{tabular} & \begin{tabular}[t]{@{}l@{}}
    person wearing black shirt and white pants \\
    person wearing white t-shirt with a peace sign \\
    person wearing a classic detective trench coat and fedora \\
    surfer wearing board shorts with a tropical pattern \\
    mountaineer in a thermal jacket and snow goggles \\
    \textbf{chef in a white jacket and checkered pants} \\
    pilot in a vintage leather jacket with aviator sunglasses \\
    \end{tabular} \\
  \hline
  \begin{tabular}[t]{@{}l@{}}
    \texttt{rp\_alexandra\_rigged\_004\_yup\_a.obj} \\
    (Renderpeople)\\
  \end{tabular} & \begin{tabular}[t]{@{}l@{}}
    person in red sweater, blue jeans \\
    person in white sweater with a red logo, yoga pants \\
    professional gamer in a team jersey and headphones \\
    ballet dancer in a pink tutu and ballet slippers \\
    rock star with leather jacket \\
    vintage 1950s dress with polka dots and sunglasses \\
    \textbf{athlete in a running outfit with a marathon number} \\
    \end{tabular} \\
  \hline
  \begin{tabular}[t]{@{}l@{}}
    \texttt{rp\_adanna\_rigged\_007\_yup\_a.obj} \\
    (Renderpeople)\\
  \end{tabular} & \begin{tabular}[t]{@{}l@{}}
    nunn in a black dress \\
    nunn in a white dress, black headscarf \\
    \textbf{professional in a suit jacket, skirt, and elegant headscarf} \\
    athlete in sportswear with a sporty hijab \\
    artist in a paint-splattered apron and a stylish hijab \\
    student in a denim jacket, casual dress, and a colorful headscarf \\
    doctor in a lab coat with a simple, modest hijab \\
    \end{tabular} \\
  \hline
  \begin{tabular}[t]{@{}l@{}}
    \texttt{rp\_aaron\_rigged\_001\_yup\_a.obj} \\
    (Renderpeople)\\
  \end{tabular} & \begin{tabular}[t]{@{}l@{}}
    railroad worker wearing high-vis vest \\
    biker wearing red jacket and black pants \\
    firefighter in full gear with reflective stripes \\
    plumber in a blue jumpsuit \\
    \textbf{electrician with a tool belt and safety goggles} \\
    carpenter in overalls with a hammer in pocket \\
    landscape gardener in a green t-shirt and cargo pants \\
    \end{tabular} \\
  \hline
  \begin{tabular}[t]{@{}l@{}}
    \texttt{human.obj} \\
    \\
  \end{tabular} & \begin{tabular}[t]{@{}l@{}}
    \textbf{a photo of spiderman} \\
    a caricature of a pirate with a large hat and eye patch \\
    a whimsical wizard with a pointed hat, dark shadow \\
    a cartoon astronaut with a bubbly space helmet \\
    a ninja turtle with a colorful mask \\
    a cartoon zombie in tattered clothes \\
    \end{tabular} \\
  \hline
\end{tabular}
\caption{Meshes and text prompts used for the user study. Most of these text prompts and meshes were also used in TexFusion \cite{cao2023texfusion}. \texttt{human.obj} is a proprietary mesh and not included in any particular dataset. Bolded prompts are shown in Table \ref{tab:sixtextures} in order of the meshes listed here.}
\label{table:prompts}
\end{table*}

\begin{table*}
\centering
\begin{tabular}{cccc}
\includegraphics[width=1.15in, trim={2cm 3cm 2cm 0}, clip,bb=0 0 724 724]{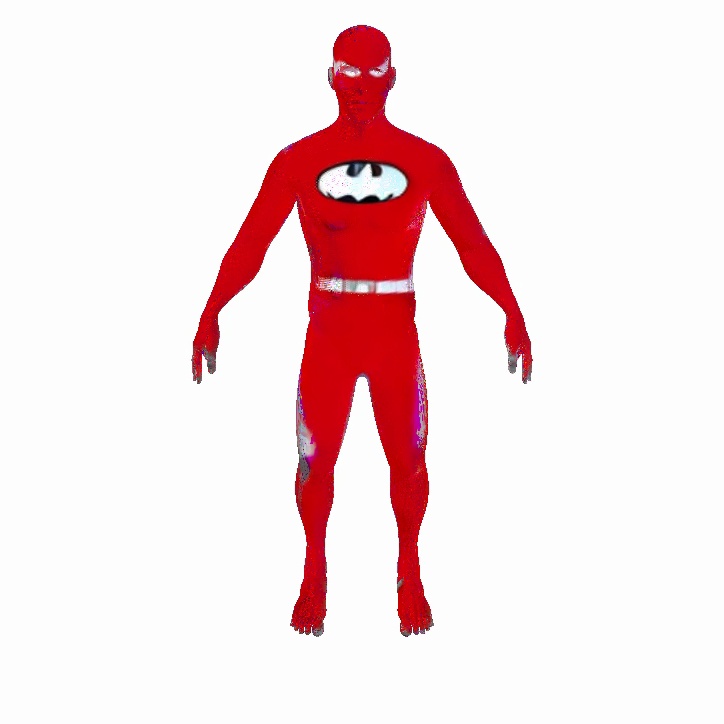} & 
\includegraphics[width=1.15in, trim={2cm 3cm 2cm 0}, clip,bb=0 0 724 724]{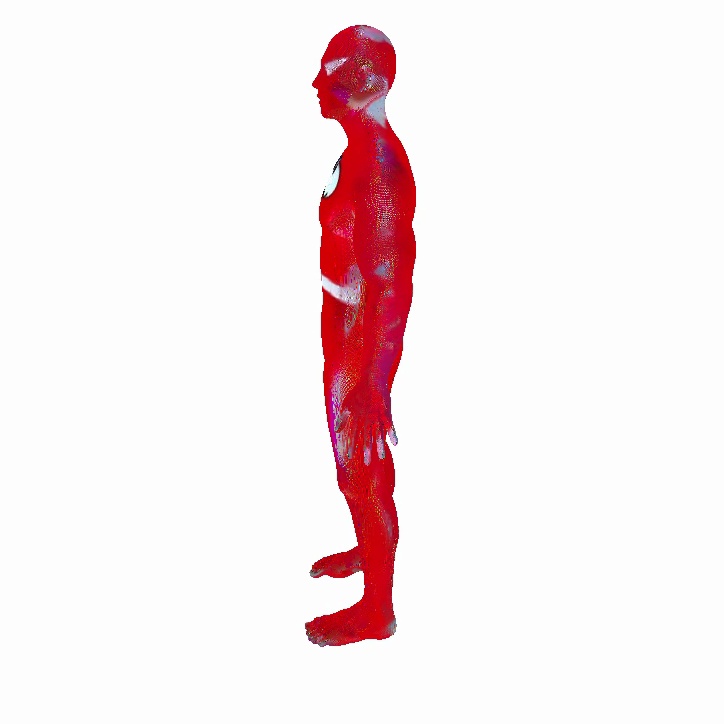} & 
\includegraphics[width=1.15in, trim={2cm 3cm 2cm 0}, clip,bb=0 0 724 724]{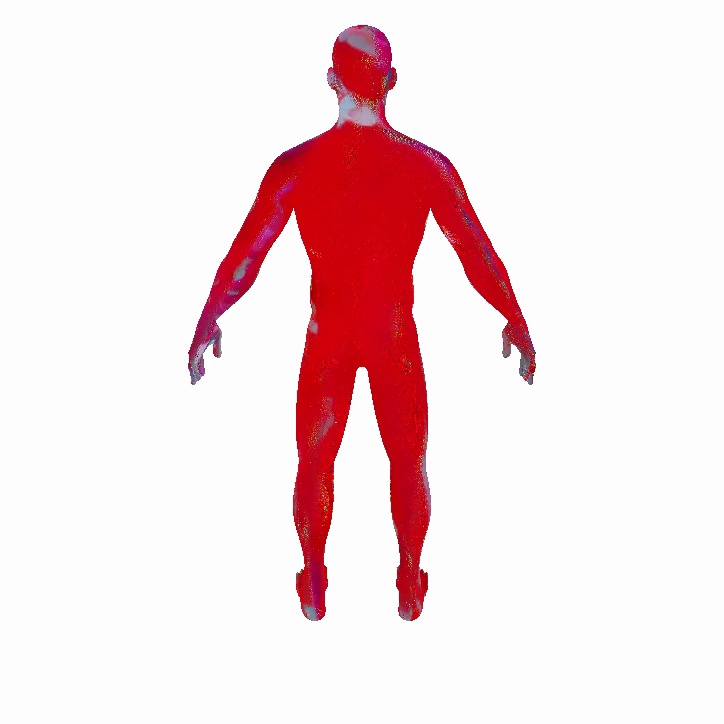} & 
\includegraphics[width=1.15in, trim={2cm 3cm 2cm 0}, clip,bb=0 0 724 724]{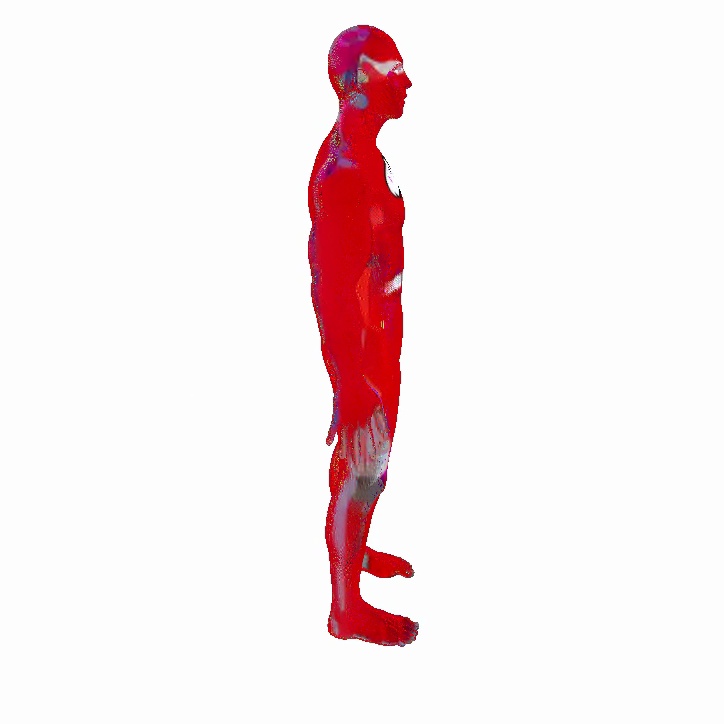} \\
\includegraphics[width=1.15in, trim={2cm 3cm 2cm 0}, clip,bb=0 0 724 724]{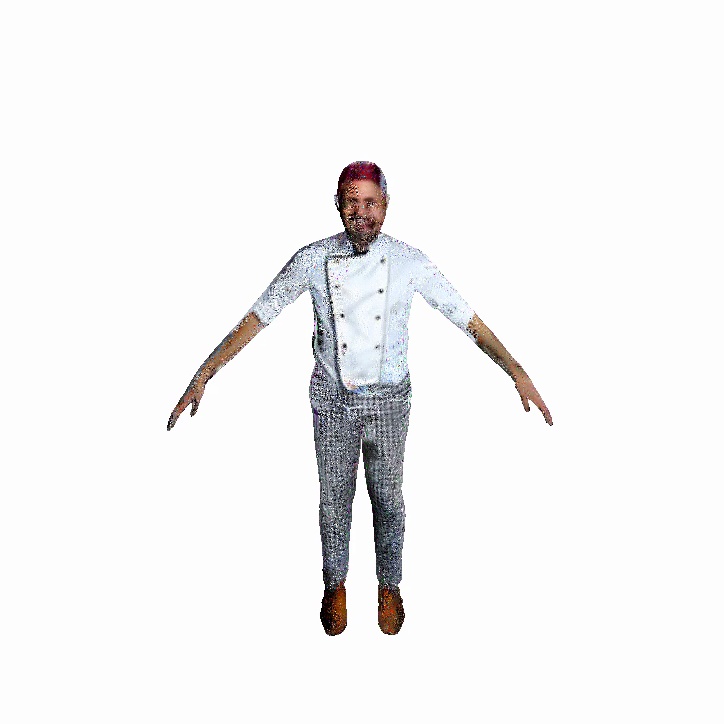} & 
\includegraphics[width=1.15in, trim={2cm 3cm 2cm 0}, clip,bb=0 0 724 724]{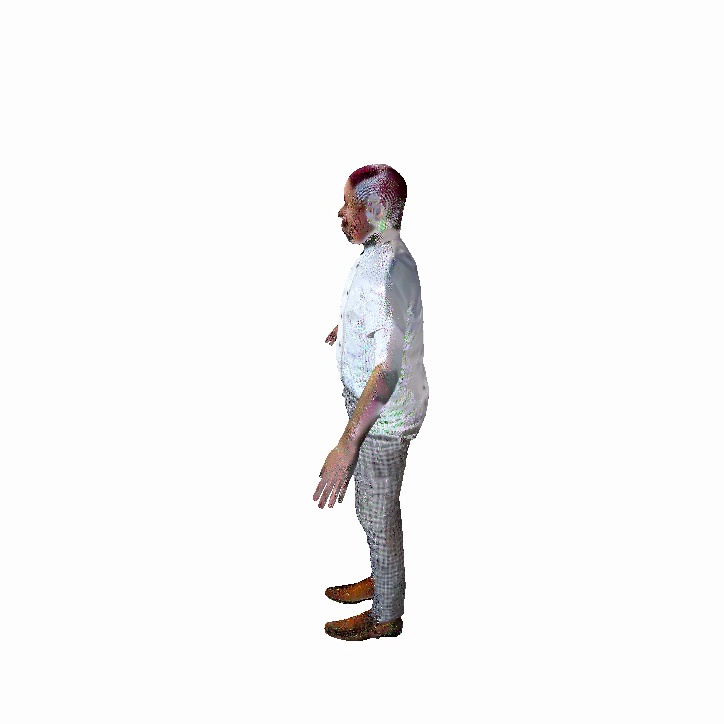} & 
\includegraphics[width=1.15in, trim={2cm 3cm 2cm 0}, clip,bb=0 0 724 724]{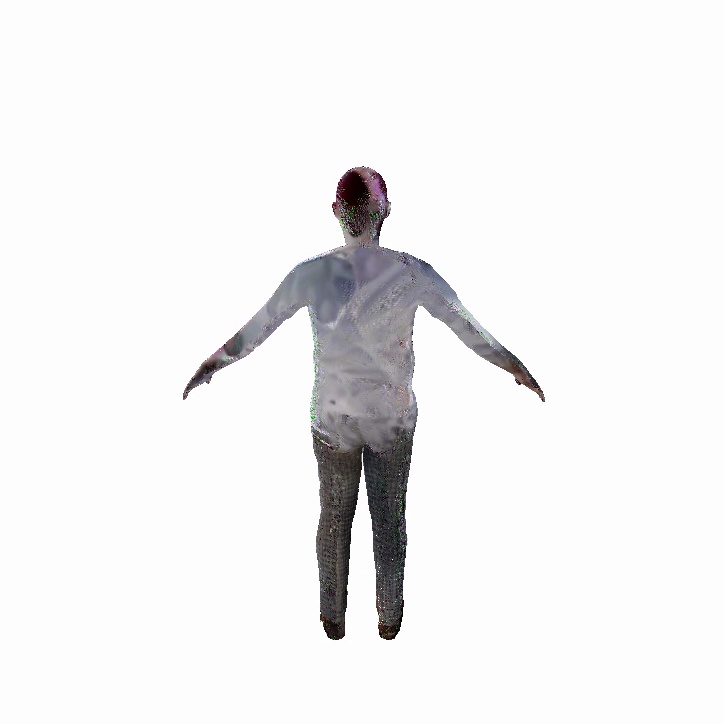} & 
\includegraphics[width=1.15in, trim={2cm 3cm 2cm 0}, clip,bb=0 0 724 724]{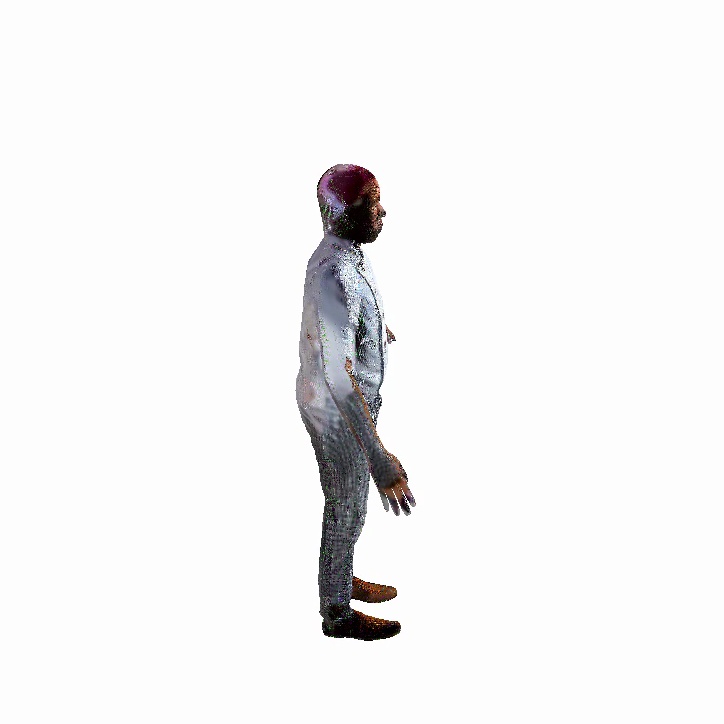} \\
\includegraphics[width=1.15in, trim={2cm 0 2cm 0}, clip,bb=0 0 724 724]{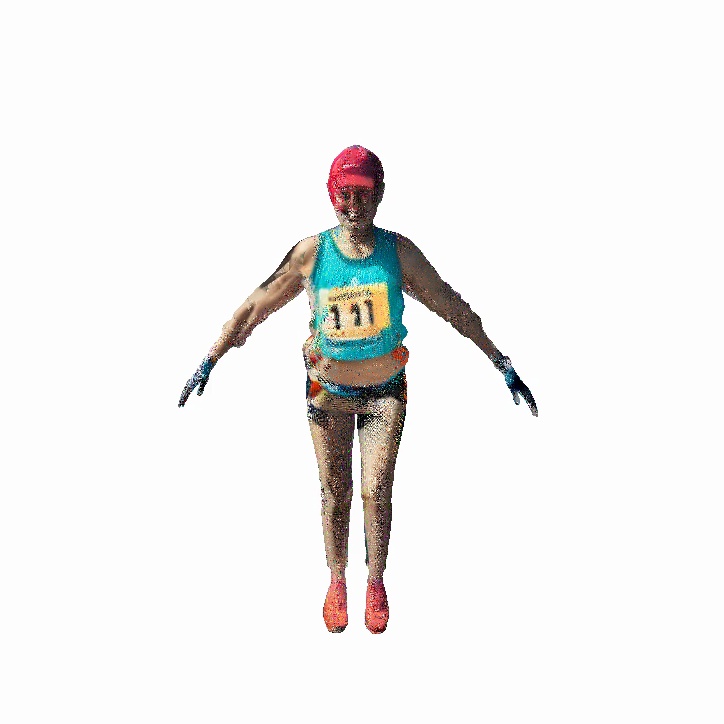} & 
\includegraphics[width=1.15in, trim={2cm 0 2cm 0}, clip,bb=0 0 724 724]{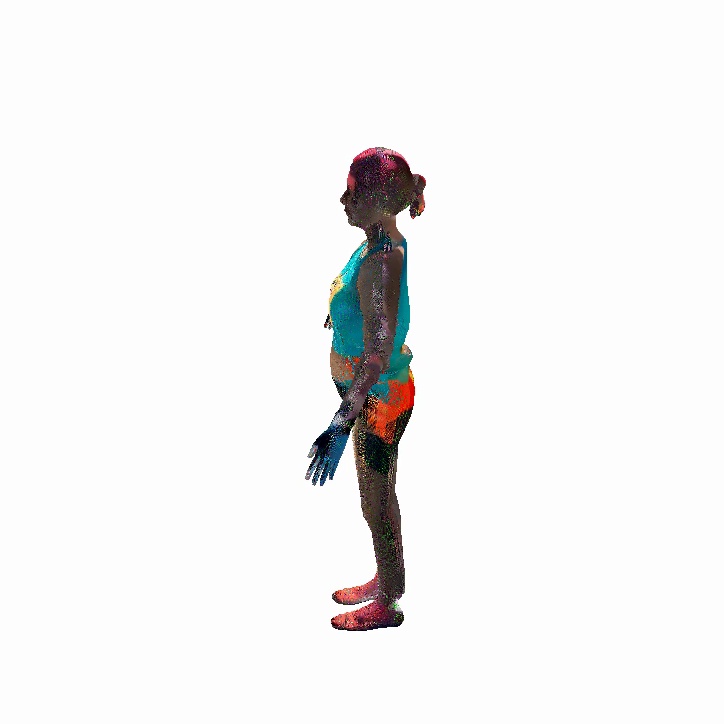} & 
\includegraphics[width=1.15in, trim={2cm 0 2cm 0}, clip,bb=0 0 724 724]{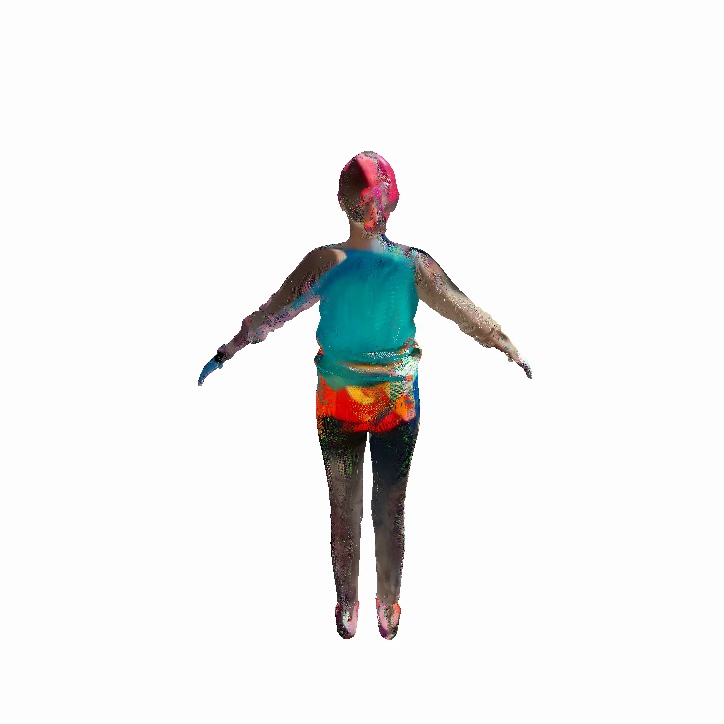} & 
\includegraphics[width=1.15in, trim={2cm 0 2cm 0}, clip,bb=0 0 724 724]{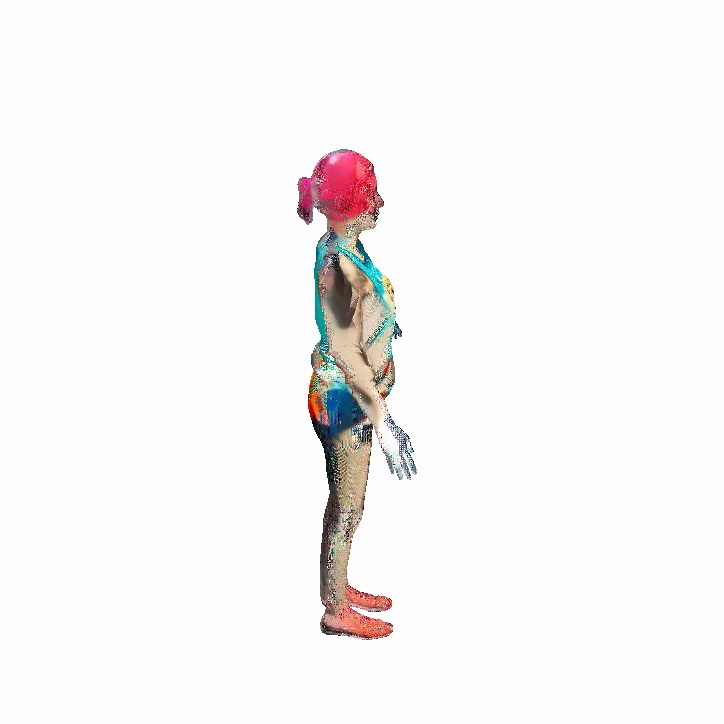} \\
\includegraphics[width=1.15in, trim={2cm 3cm 2cm 0}, clip,bb=0 0 724 724]{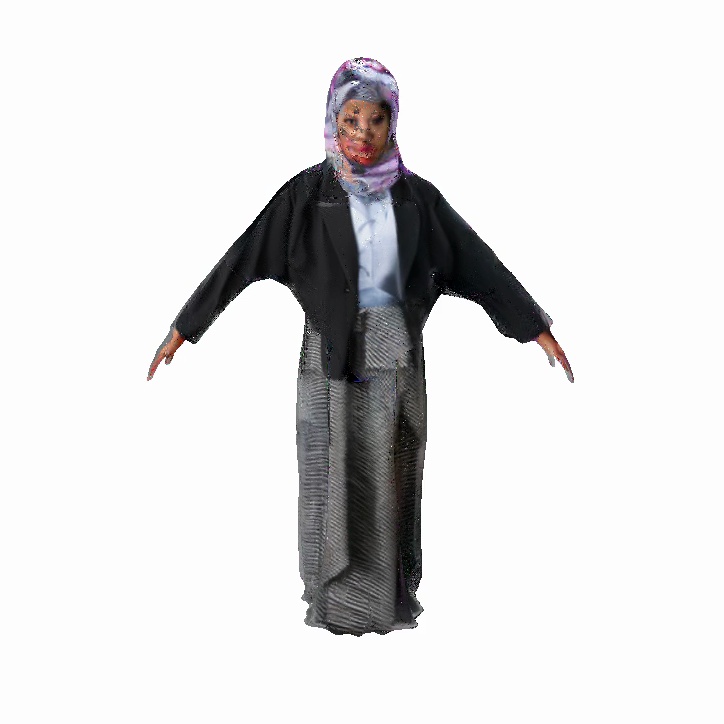} & 
\includegraphics[width=1.15in, trim={2cm 3cm 2cm 0}, clip,bb=0 0 724 724]{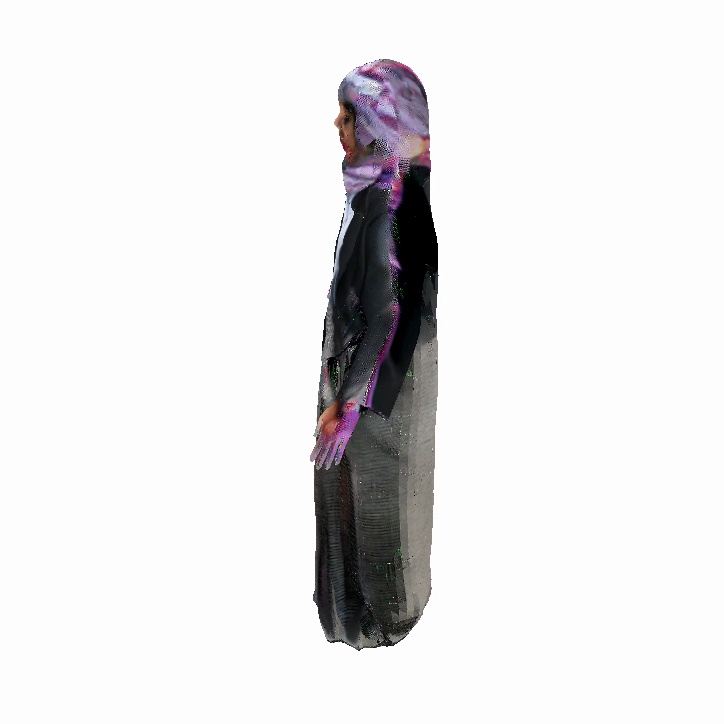} & 
\includegraphics[width=1.15in, trim={2cm 3cm 2cm 0}, clip,bb=0 0 724 724]{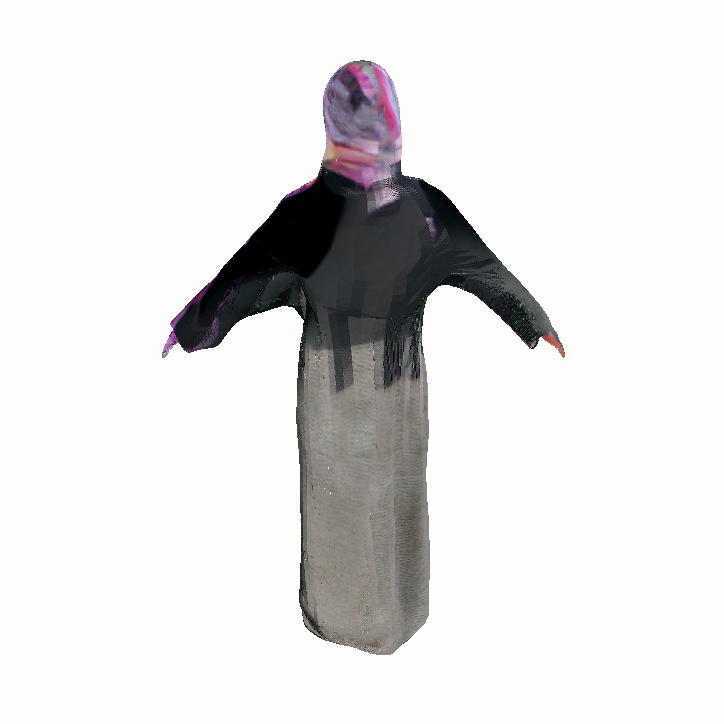} & 
\includegraphics[width=1.15in, trim={2cm 3cm 2cm 0}, clip,bb=0 0 724 724]{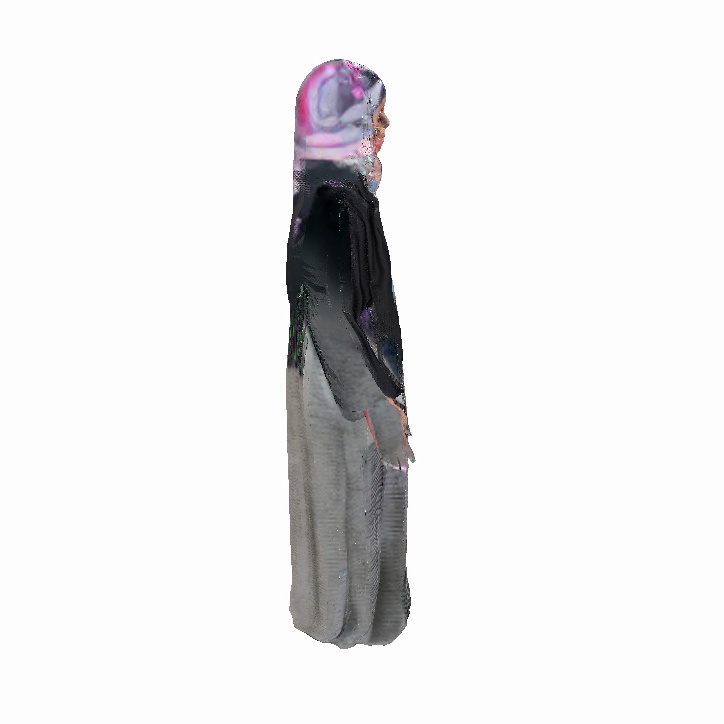} \\
\includegraphics[width=1.15in, trim={2cm 0 2cm 0}, clip,bb=0 0 724 724]{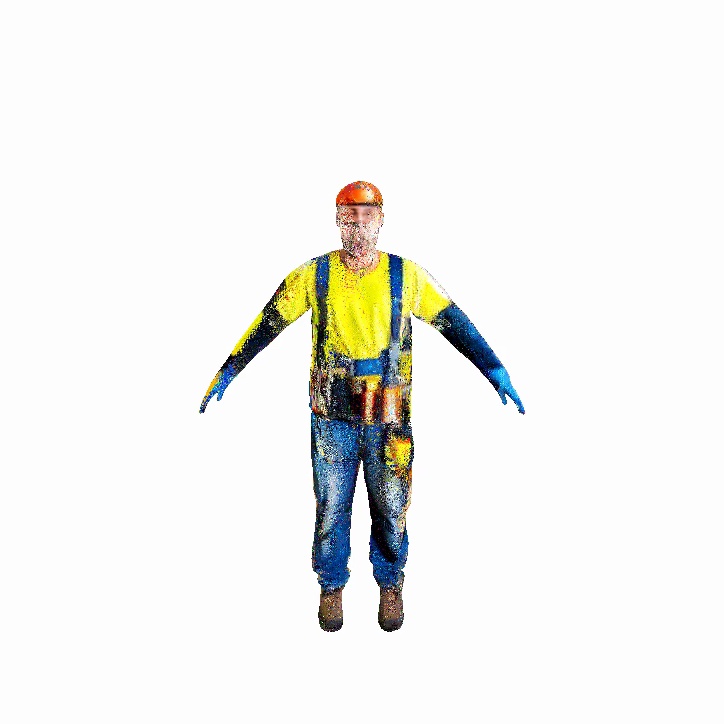} & 
\includegraphics[width=1.15in, trim={2cm 0 2cm 0}, clip,bb=0 0 724 724]{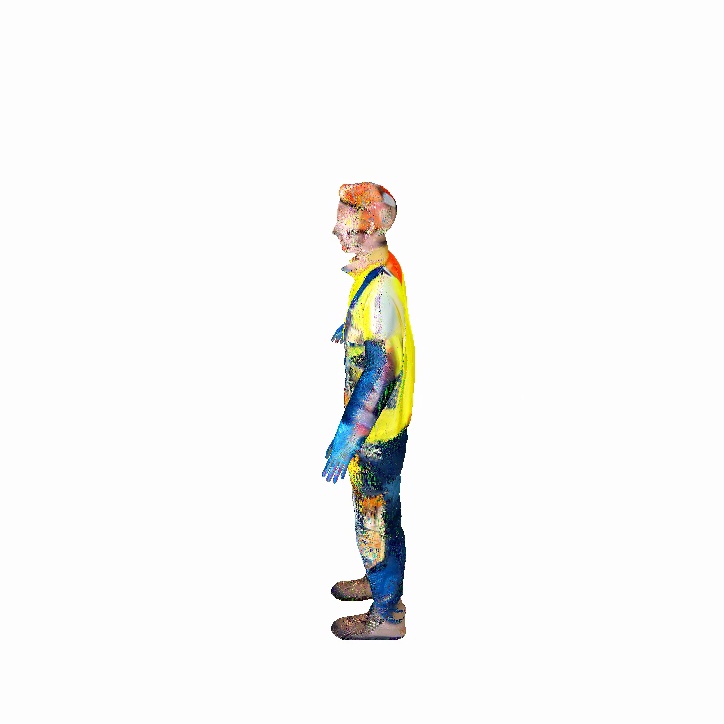} & 
\includegraphics[width=1.15in, trim={2cm 0 2cm 0}, clip,bb=0 0 724 724]{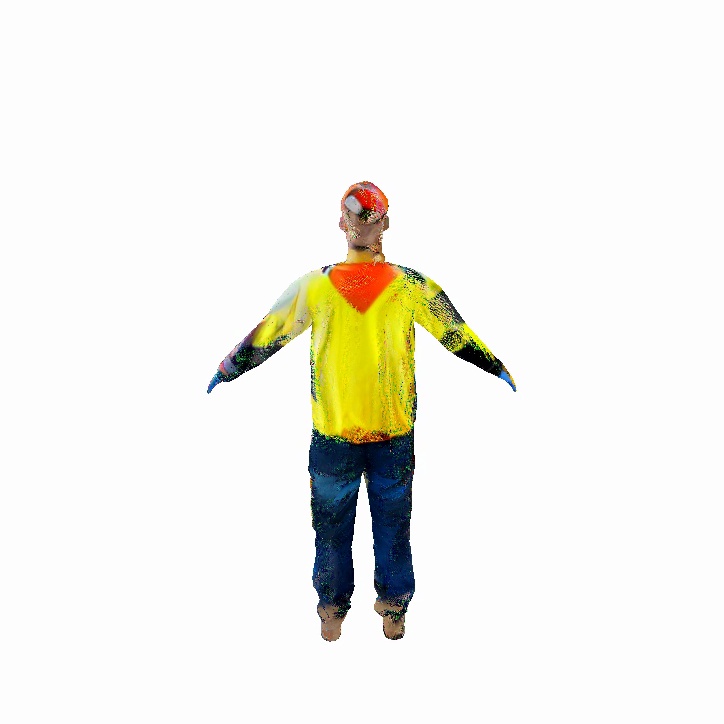} & 
\includegraphics[width=1.15in, trim={2cm 0 2cm 0}, clip,bb=0 0 724 724]{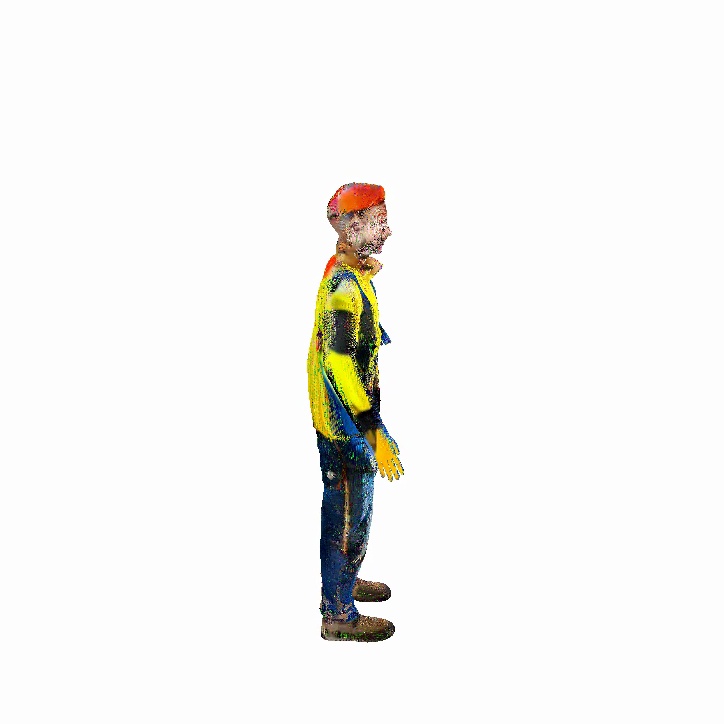} \\
\includegraphics[width=1.15in, trim={2cm 0 2cm 0}, clip,bb=0 0 724 724]{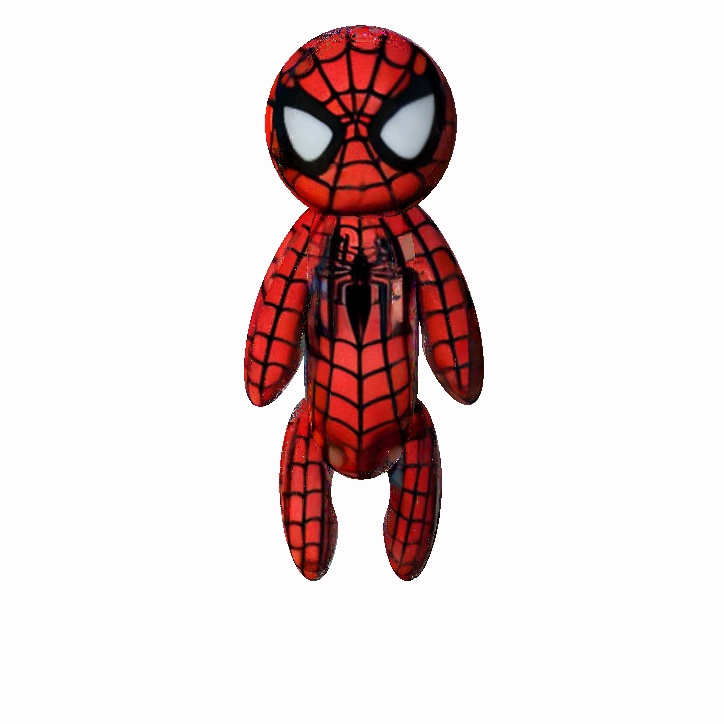} & 
\includegraphics[width=1.15in, trim={2cm 0 2cm 0}, clip,bb=0 0 724 724]{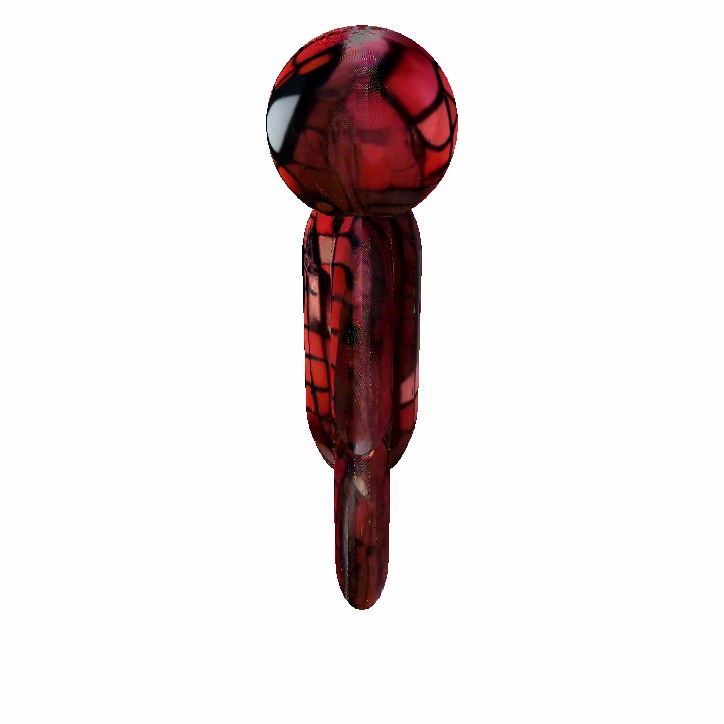} & 
\includegraphics[width=1.15in, trim={2cm 0 2cm 0}, clip,bb=0 0 724 724]{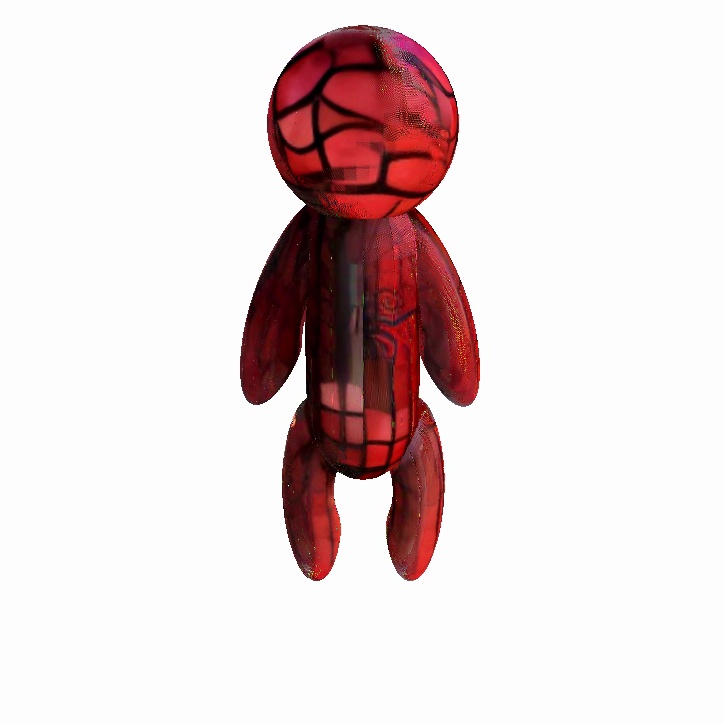} & 
\includegraphics[width=1.15in, trim={2cm 0 2cm 0}, clip,bb=0 0 724 724]{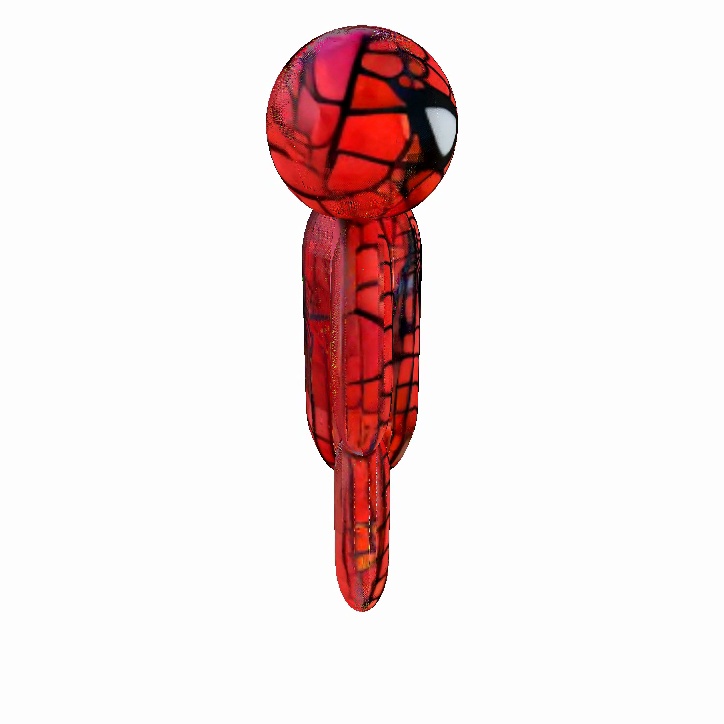} \\
\end{tabular}
\caption{Six texture maps constructed using ConTEXTure. Each image on each row was rendered from the texture map from four equidistant azimuth angles, $0\degree$, $90\degree$, $180\degree$, and $270\degree$ (from left to right).}
\label{tab:sixtextures}
\end{table*}

\clearpage